\documentclass[runningheads]{llncs}
\usepackage{graphicx}
\usepackage{comment}
\usepackage{amsmath,amssymb} 
\usepackage{color}
\usepackage{xcolor}
\usepackage[british,english,american]{babel}
\usepackage{xspace}
\usepackage{enumitem}
\usepackage{subcaption}
\usepackage{multirow}
\usepackage{array}

\makeatletter
\DeclareRobustCommand\onedot{\futurelet\@let@token\@onedot}
\def\@onedot{\ifx\@let@token.\else.\null\fi\xspace}

\def\eg{\emph{e.g}\onedot} 
\def\ie{\emph{i.e}\onedot} 
 
 \def\vs{\emph{vs}\onedot}

\makeatother

\newcommand{\app}{\raise.17ex\hbox{$\scriptstyle\sim$}}

\newcommand{\unnas}{{UnNAS}}
\newcommand{\unnasdarts}{{UnNAS-\texttt{DARTS}}}
\newcommand{\nasdarts}{{NAS-\texttt{DARTS}}}
\newcommand{\cls}{{\texttt{Supv.Cls}}}
\newcommand{\seg}{{\texttt{Supv.Seg}}}
\newcommand{\jig}{{\texttt{Jigsaw}}}
\newcommand{\col}{{\texttt{Color}}}
\newcommand{\rot}{{\texttt{Rot}}}

\newcommand{\pmstd}[1]{$_{\pm\text{#1}}$}
\newcommand{\ft}[1]{\fontsize{#1pt}{1em}\selectfont}
\newcommand{\grayfy}[1]{{\textcolor{gray}{#1}}}

\newlength\savewidth\newcommand\shline{\noalign{\global\savewidth\arrayrulewidth
  \global\arrayrulewidth 1pt}\hline\noalign{\global\arrayrulewidth\savewidth}}
\newcommand{\tablestyle}[2]{\setlength{\tabcolsep}{#1}\renewcommand{\arraystretch}{#2}\centering\footnotesize}

\begin{document}
\pagestyle{headings}
\mainmatter
\def\ECCVSubNumber{2767}  

\title{Are Labels Necessary for \\Neural Architecture Search?
} 

\titlerunning{Are Labels Necessary for Neural Architecture Search?}
%
\author{Chenxi Liu\inst{1} \and
Piotr Doll\'ar\inst{2} \and
Kaiming He\inst{2} \and
Ross Girshick\inst{2} \and
Alan Yuille\inst{1} \and
Saining Xie\inst{2}
}
\authorrunning{C. Liu et al.}
%
\institute{Johns Hopkins University \and
Facebook AI Research}
\maketitle

\begin{abstract} 
Existing neural network architectures in computer vision --- whether designed by humans or by machines --- were typically found using both images and their associated labels. In this paper, we ask the question: can we find high-quality neural architectures using only images, but no human-annotated labels? To answer this question, we first define a new setup called Unsupervised Neural Architecture Search (UnNAS). We then conduct two sets of experiments. In sample-based experiments, we train a large number (500) of diverse architectures with either supervised or unsupervised objectives, and find that the architecture rankings produced with and without labels are highly correlated. In search-based experiments, we run a well-established NAS algorithm (DARTS) using various unsupervised objectives, and report that the architectures searched without labels can be competitive to their counterparts searched with labels. Together, these results reveal the potentially surprising finding that labels are not necessary, and the image statistics alone may be sufficient to identify good neural architectures.\footnote{Code release: \url{https://github.com/facebookresearch/unnas}}
\end{abstract}


\section{Introduction}

Neural architecture search (NAS) has emerged as a research problem of searching for architectures that perform well on target data and tasks. A key mystery surrounding NAS is what factors contribute to the success of the search. Intuitively, using the target data and tasks during the search will result in the least domain gap, and this is indeed the strategy adopted in early NAS attempts~\cite{Zoph2017,real2017large}. Later, researchers~\cite{zoph2018learning} started to utilize the transferability of architectures, which enabled the search to be performed on different data and labels (\eg, CIFAR-10) than the target (\eg, ImageNet). However, what has not changed is that \emph{both the images and the (semantic) labels} provided in the dataset need to be used in order to search for an architecture. In other words, existing NAS approaches perform search in the \emph{supervised} learning regime. 

In this paper, we take a step towards understanding what role supervision plays in the success of NAS. We ask the question: How indispensable are labels in neural architecture search? Is it possible to find high-quality architectures using images only? This corresponds to the important yet underexplored \emph{unsupervised setup} of neural architecture search, which we formalize in Section~\ref{sec:unnas}. 

With the absence of labels, the quality of the architecture needs to be estimated in an unsupervised fashion during the search phase. In the present work, we conduct two sets of experiments using three unsupervised training methods \cite{gidaris2018unsupervised,zhang2016colorful,noroozi2016unsupervised} from the recent self-supervised learning literature.\footnote{Self-supervised learning is a form of unsupervised learning. The term ``unsupervised learning'' in general emphasizes ``without \emph{human}-annotated labels'', while the term ``self-supervised learning'' emphasizes ``producing labels from data''.} These two sets of experiments approach the question from complementary perspectives. In \emph{sample-based experiments}, we randomly sample 500 architectures from a search space, train and evaluate them using supervised \vs self-supervised objectives, and then examine the rank correlation (when sorting models by accuracy) between the two training methodologies. In \emph{search-based experiments}, we take a well-established NAS algorithm, replace the supervised search objective with a self-supervised one, and examine the quality of the searched architecture on tasks such as ImageNet classification and Cityscapes semantic segmentation. Our findings include:

\begin{itemize}[itemsep=.1em,topsep=.5em]
\item The architecture rankings produced by supervised and self-supervised pretext tasks are \emph{highly correlated}. This finding is consistent across two datasets, two search spaces, and three pretext tasks.
\item The architectures searched without human annotations are \emph{comparable in performance} to their supervised counterparts. This result is consistent across three pretext tasks, three pretext datasets, and two target tasks. There are even cases where unsupervised search outperforms supervised search.
\item Existing NAS approaches typically use \emph{labeled} images from a \emph{smaller} dataset to learn transferable architectures. We present evidence that using \emph{unlabeled} images from a \emph{large} dataset may be a more promising approach. 
\end{itemize}

We conclude that labels are not necessary for neural architecture search, and the deciding factor for architecture quality may hide within the image pixels. 

\section{Related Work}

\subsubsection{Neural Architecture Search.} Research on the NAS problem involves designing the search space \cite{zoph2018learning,ying2019bench} and the search algorithm \cite{Zoph2017,real2019regularized}. There are special focuses on reducing the overall time cost of the search process \cite{liu2018progressive,pham2018efficient,liu2018darts}, or on extending to a larger variety of tasks \cite{chen2018searching,liu2019auto,ghiasi2019fpn,so2019evolved}. Existing works on NAS all use human-annotated labels during the search phase. Our work is orthogonal to existing NAS research, in that we explore the unsupervised setup.

\subsubsection{Architecture Transferability.} In early NAS attempts \cite{Zoph2017,real2017large}, the search phase and the evaluation phase typically operate on the same dataset and task. Later, researchers realized that it is possible to relax this constraint. In these situations, the dataset and task used in the search phase are typically referred as the \emph{proxy} to the target dataset and task, reflecting a notion of architecture transferability. \cite{zoph2018learning} demonstrated that CIFAR-10 classification is a good proxy for ImageNet classification. \cite{liu2018progressive} measured the rank correlation between these two tasks using a small number of architectures. \cite{kornblith2019better} studied the transferability of 16 architectures (together with trained weights) between more \emph{supervised} tasks. 
\cite{kolesnikov2019revisiting} studied the role of architecture in several self-supervised tasks, but with a small number of architectures and a different evaluation method (\ie linear probe). 
Part of our work studies architecture transferability at a larger scale, \emph{across} supervised and unsupervised tasks. 

\subsubsection{Unsupervised Learning.} There is a large literature on unsupervised learning, \eg, \cite{Dosovitskiy2014,Doersch2015,Wang2015,zhang2016colorful,noroozi2016unsupervised,gidaris2018unsupervised,Oord2018,Wu2018a,Tian2019,he2019momentum}. In the existing literature, methods are generally developed to learn the \emph{weights} (parameters) of a fixed architecture without using labels, and these weights are evaluated by transferring to a target \emph{supervised} task. In our study, we explore the possibility of using such methods to learn the \emph{architecture} without using labels, rather than the weights. 
Therefore, our subject of study is simultaneously the unsupervised generalization of NAS, and the architecture level generalization of unsupervised learning. 

\section{Unsupervised Neural Architecture Search}
\label{sec:unnas}

The goal of this paper is to provide an answer to the question asked in the title: are labels necessary for neural architecture search? To formalize this question, in this section, we define a new setup called Unsupervised Neural Architecture Search (\unnas{}). Note that UnNAS represents a general problem setup instead of any specific algorithm for solving this problem. We instantiate \unnas{} with specific algorithms and experiments to explore the importance of labels in neural architecture search.

\subsection{Search Phase}
The traditional NAS problem includes a search phase: given a pre-defined search space, the search algorithm explores this space and estimates the performance (\eg accuracy) of the architectures sampled from the space. The accuracy estimation can involve full or partial training of an architecture. Estimating the accuracy requires access to the labels of the dataset. So the traditional NAS problem is essentially a \emph{supervised learning} problem.

We define UnNAS as the counterpart \emph{unsupervised learning} problem. It follows the definition of the NAS problem, only except that there are no human-annotated labels provided for estimating the performance of the architectures. An algorithm for the UnNAS problem still explores a pre-defined search space,\footnote{Admittedly, the search space design is typically heavily influenced by years of manual search, usually with supervision.} but it requires other criteria to estimate how good a sampled architecture is. 

\subsection{Evaluation Phase}

Generally speaking, the goal of the NAS problem is to find an architecture. The weights of the found architecture are \emph{not} necessarily the output of a NAS algorithm. Instead, the weights are optimized after the search phase in an evaluation phase of the NAS problem: it includes training the found architecture on a target dataset's training split, and validating the accuracy on the target dataset's validation split. We note that ``\emph{training} the architecture weights" is part of the NAS \emph{evaluation} phase---the labels in the target dataset, both training and validation splits, play a role of evaluating an architecture.

Based on this context, we define \emph{the evaluation phase of \unnas{}} in the same way: training the weights of the architecture (found by an \unnas{} algorithm) on a target dataset's training split, and validating the accuracy on the target dataset's validation split, both \emph{using} the labels of the target dataset. We remark that using labels during the evaluation phase does \emph{not} conflict with the definition of \unnas{}: the search phase is unsupervised, while the evaluation phase requires labels to examine how good the architecture is.

\begin{figure}[t]\centering
\includegraphics[width=0.55\linewidth]{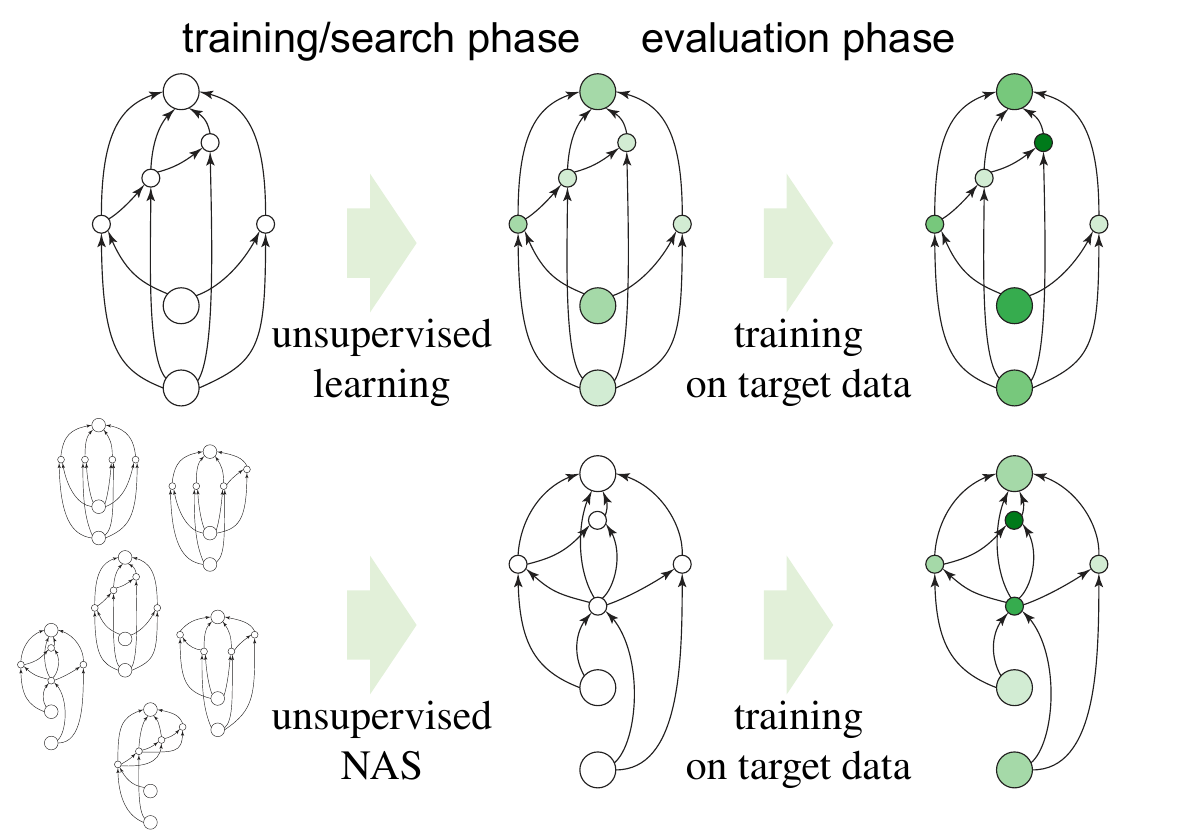}
  \caption{\textbf{Unsupervised neural architecture search}, or \mbox{UnNAS}, is a new problem setup that helps answer the question: are labels necessary for neural architecture search? In traditional unsupervised learning (top panel), the \emph{training phase} learns the weights of a fixed architecture; then the \emph{evaluation phase} measures the quality of the weights by training a classifier (either by fine-tuning the weights or using them as a fixed feature extractor) using supervision from the target dataset. Analogously, in \mbox{UnNAS} (bottom panel), the \emph{search phase} searches for an architecture without using labels; and the \emph{evaluation phase} measures the quality of the architecture found by an \mbox{UnNAS} algorithm by training the architecture's weights using supervision from the target dataset.
}
\label{fig:setup}
\end{figure}

\subsection{Analogy to Unsupervised Learning}

Our definition of \unnas{} is analogous to unsupervised \emph{weight} learning in existing literature \cite{gidaris2018unsupervised,zhang2016colorful,noroozi2016unsupervised}. While the unsupervised learning phase has no labels for training weights, the quality of the learned weights is evaluated by transferring them to a target task, supervised by labels. We emphasize that in the \unnas{} setup, labels play an analogous role during evaluation. See Figure~\ref{fig:setup} for an illustration and elaboration on this analogy.

Similar to unsupervised weight learning, in principle, the search dataset should be different than the evaluation (target) dataset in \unnas{} in order to more accurately reflect real application scenarios.

\section{Experiments Overview}

As Section~\ref{sec:unnas} describes, an architecture discovered in an \emph{unsupervised} fashion will be evaluated by its performance in a \emph{supervised} setting. Therefore, we are essentially looking for some type of architecture level \emph{correlation} that can reach across the unsupervised \vs supervised boundary, so that the unsupervisedly discovered architecture could be reliably \emph{transferred} to the supervised target task. We investigate whether several existing self-supervised pretext tasks (described in Section~\ref{sec:pretext}) can serve this purpose, through two sets of experiments of complementary nature: \emph{sample-based} (Section~\ref{sec:sample-based}) and \emph{search-based} (Section~\ref{sec:search-based}). In \emph{sample-based}, each network is trained and evaluated individually, but the downside is that we can only consider a small, random subset of the search space. In \emph{search-based}, the focus is to find a top architecture from the entire search space, but the downside is that the training dynamics during the search phase does not exactly match that of the evaluation phase. 

\subsection{Pretext Tasks}
\label{sec:pretext}

We explore three unsupervised training methods (typically referred to as \emph{pretext tasks} in self-supervised learning literature): rotation prediction, colorization, and solving jigsaw puzzles. We briefly describe them for completeness.

\begin{itemize}[itemsep=.1em,topsep=.5em]
\item \emph{Rotation prediction}~\cite{gidaris2018unsupervised} (\rot): the input image undergoes one of four preset rotations (0, 90, 180, and 270 degrees), and the pretext task is formulated as a 4-way classification problem that predicts the rotation.

\item \emph{Colorization}~\cite{zhang2016colorful} (\col): the input is a grayscale image, and the pretext task is formulated as a pixel-wise classification problem with a set of pre-defined color classes (313 in \cite{zhang2016colorful}). 

\item \emph{Solving jigsaw puzzles}~\cite{noroozi2016unsupervised} (\jig): the input image is divided into patches and randomly shuffled. The pretext task is formulated as an image-wise classification problem that chooses from one out of $K$ preset permutations.\footnote{On ImageNet, each image is divided into 3$\times$3$=$9 patches and $K$ is 1000 selected from 9!$=$362,880 permutations \cite{noroozi2016unsupervised}; on CIFAR-10 in which images are smaller, we use 2$\times$2$=$4 patches and $K$ is 4!$=$24 permutations.}
\end{itemize}

All three pretext tasks are image or pixel-wise classification problems. Therefore, we can compute the classification accuracy of the pretext task (on a validation set). Based on this pretext task accuracy, we can analyze its correlation with the supervised classification accuracy (also on a validation set), as in our sample-based experiments (Section~\ref{sec:sample-based}). Also, since these pretext tasks all use cross entropy loss, it is also straightforward to use them as the training objective in standard NAS algorithms, as done in our search-based experiments (Section~\ref{sec:search-based}). 

\section{Sample-Based Experiments}
\label{sec:sample-based}

\begin{figure}[t]
\centering
\includegraphics[width=0.85\linewidth]
{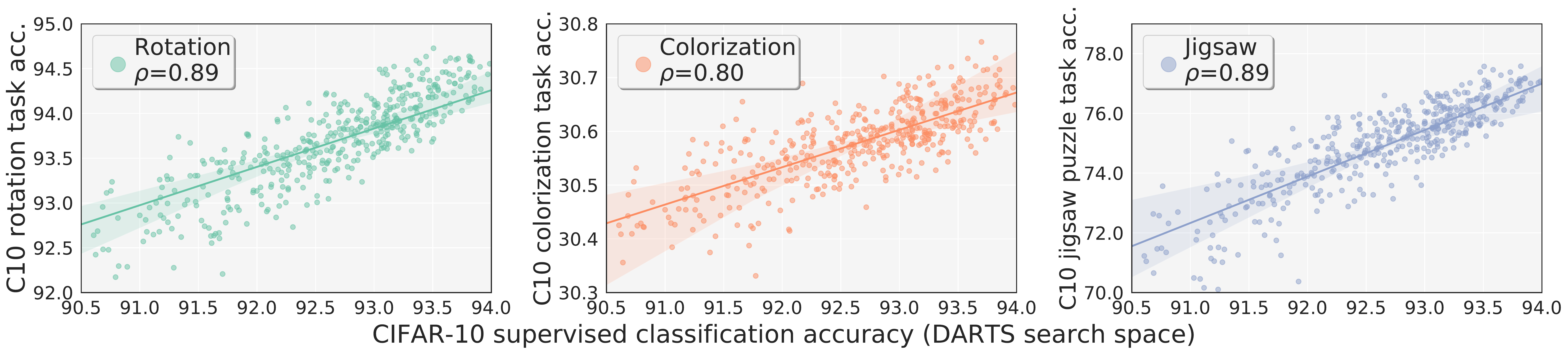}
\includegraphics[width=0.85\linewidth]
{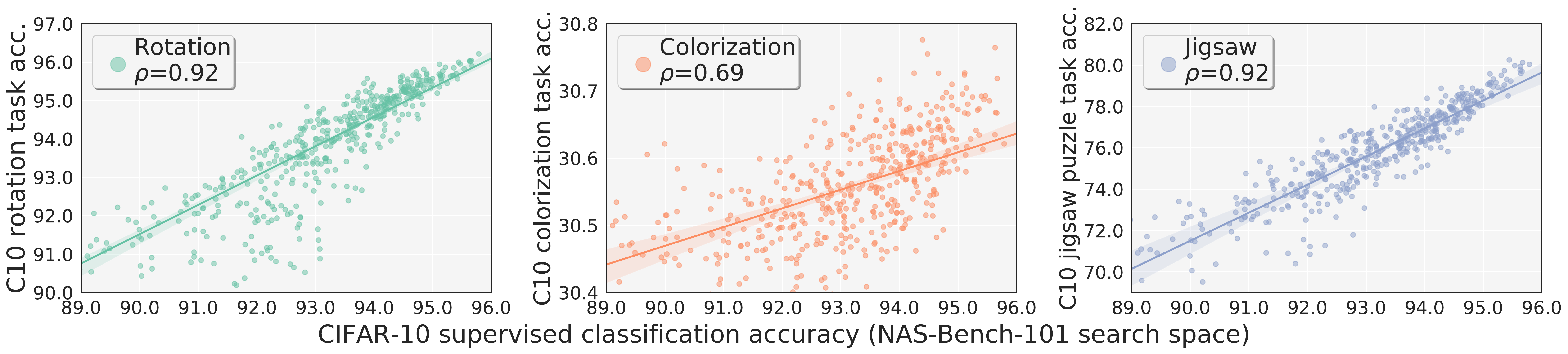}
  \caption{\textbf{Correlation between supervised classification accuracy \vs pretext task accuracy on CIFAR-10 (``C10'')}. Top panel: DARTS search space. Bottom panel: NAS-Bench-101 search space. The straight lines are fit with robust linear regression \cite{Huber2011} (same for Figure~\ref{fig:RC_IN1K} and Figure~\ref{fig:RC_across}).}
\label{fig:RC_C10}
\end{figure}

\subsection{Experimental Design}

In \emph{sample-based experiments}, we first randomly sample 500 architectures from a certain search space. We train each architecture from scratch on the pretext task and get its pretext task accuracy (\eg, the 4-way classification accuracy in the rotation task), and also train the same architecture from scratch on the supervised classification task (\eg, 1000-way classification on ImageNet).\footnote{These architectures only have small, necessary differences when trained towards these different tasks (\ie, having 4 neurons or 1000 neurons at the output layer).}

With these data collected, we perform two types of analysis.
For the \emph{rank correlation} analysis, we empirically study the statistical rank correlations between the pretext task accuracy and the supervised accuracy, by measuring the Spearman's Rank Correlation \cite{spearman1904proof}, denoted as $\rho$. 
For the \emph{random experiment} analysis, we follow the setup proposed in \cite{bergstra2012random} and recently adopted in \cite{radosavovic2019network}. Specifically, for each experiment size $m$, we sample $m$ architectures from our pool of $n$ $=$ 500 architectures. For each pretext task, we select the architecture with the highest pretext task accuracy among the $m$. This process is repeated $\lceil n/m \rceil$ times, to compute the mean and error bands ($\pm$ 2 standard deviations) of the top-1 accuracy of these $\lceil n/m \rceil$ architectures \emph{on the target dataset/task}. 

These two studies provide complementary views. The \emph{rank correlation} analysis aims to provide a global picture for \emph{all} architectures sampled from a search space, while the \emph{random experiment} analysis focuses on the \emph{top} architectures in a random experiment of varying size.

We study two search spaces: the {{DARTS search space}} \cite{liu2018darts}, and the {{NAS-Bench-101 search space}} \cite{ying2019bench}. The latter search space was built for benchmarking NAS algorithms, so we expect it to be less biased towards the search space for a certain algorithm. The experiments are conducted on two commonly used datasets: CIFAR-10~\cite{Krizhevsky2009} and ImageNet~\cite{Deng2009}.

\subsection{Implementation Details}

Details of sampling the 500 architectures are described in Appendix.
In the DARTS search space, regardless of the task, each network is trained for 5 epochs with width 32 and depth 22 on ImageNet; 100 epochs with width 16 and depth 20 on CIFAR-10.
In the NAS-Bench-101 search space, regardless of the task, each network is trained for 10 epochs with width 128 and depth 12 on ImageNet; 100 epochs with width 128 and depth 9 on CIFAR-10. Please refer to~\cite{liu2018darts,ying2019bench} for the respective definitions of \emph{width} and \emph{depth}. The performance of each network on CIFAR-10 is the average of 3 independent runs to reduce variance. 

We remark that the small-scale of CIFAR-10 dataset and the short training epochs on ImageNet are the compromises we make to allow for more diverse architectures (\ie 500). 

\begin{figure}[t]
\centering
\includegraphics[width=0.85\linewidth]
{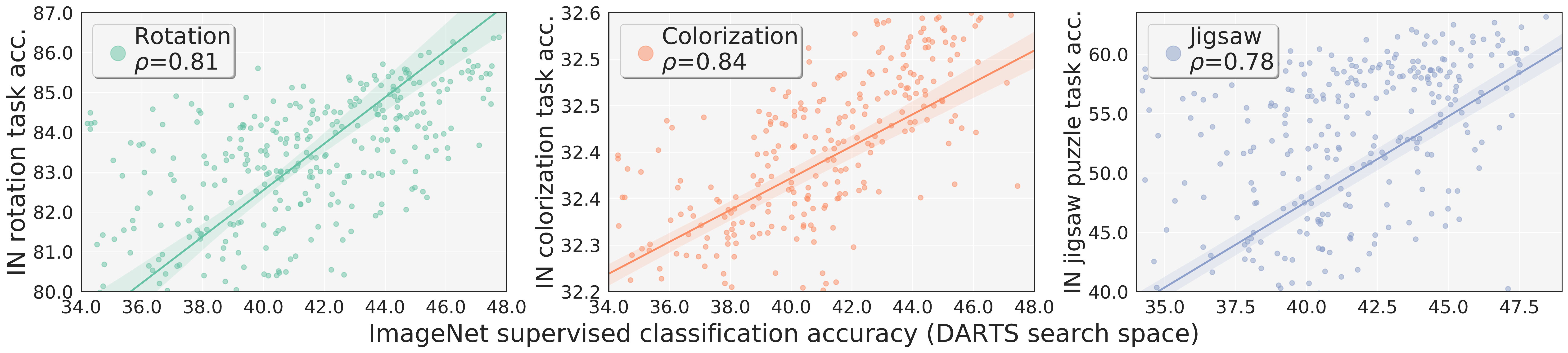}
\includegraphics[width=0.85\linewidth]
{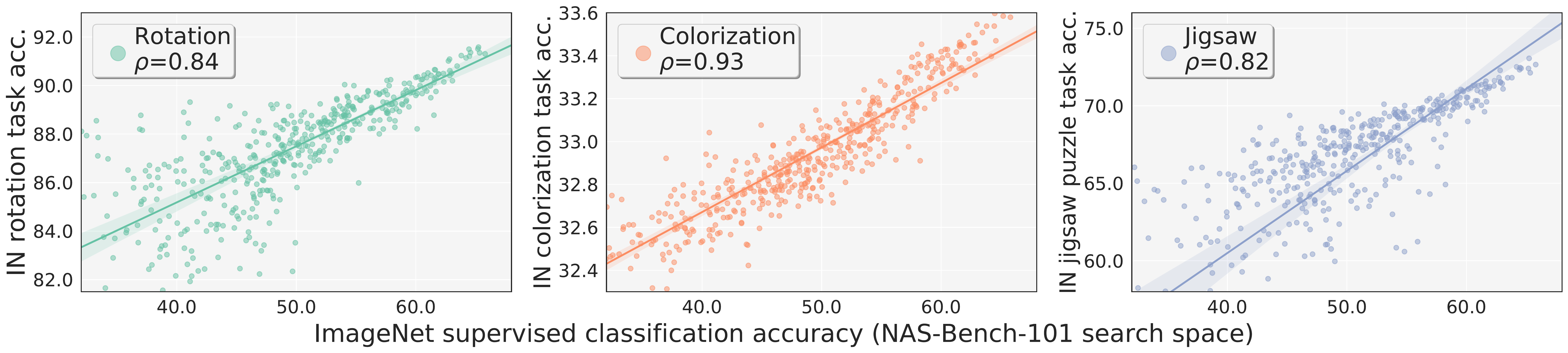}
\caption{\textbf{Correlation between supervised classification accuracy \vs pretext task accuracy on ImageNet (``IN'')}. Top panel: DARTS search space. Bottom panel: NAS-Bench-101 search space.}
\label{fig:RC_IN1K}
\end{figure}

\subsection{Results}

\subsubsection{High rank correlation between supervised accuracy and pretext accuracy on the \underline{same} dataset.}

\begin{figure}[t]
\centering
\includegraphics[width=\linewidth]{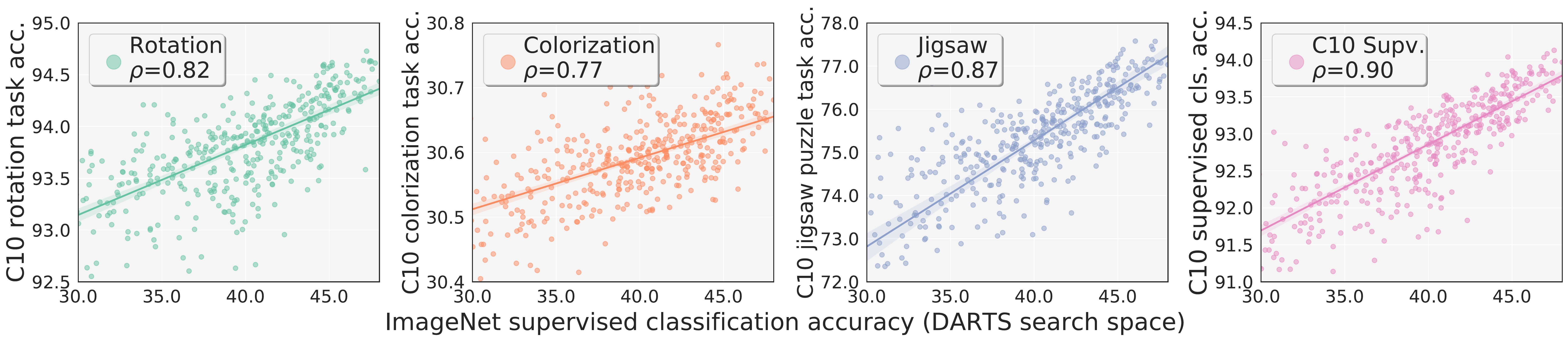}
\includegraphics[width=\linewidth]{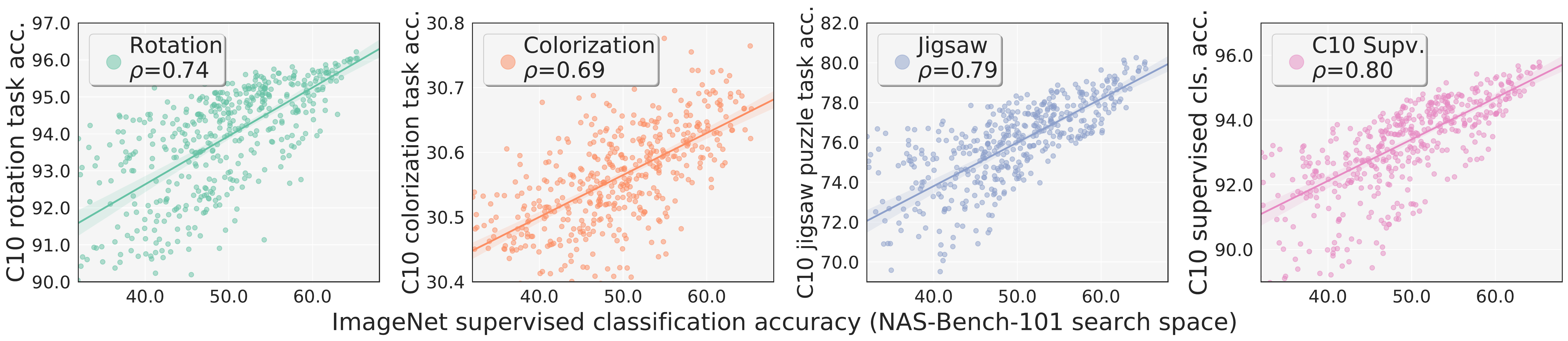}
  \caption{\textbf{Correlation between ImageNet supervised classification accuracy \vs CIFAR-10 (``C10'') pretext task accuracy}. Rankings of architectures are highly correlated between supervised classification and three unsupervised tasks, as measured by Spearman's rank correlation ($\rho$). We also show rank correlation using CIFAR-10 supervised proxy in the rightmost panel. 
Top panel: DARTS search space. Bottom panel: NAS-Bench-101 search space. }
\label{fig:RC_across}
\end{figure}

In Figure~\ref{fig:RC_C10} and Figure~\ref{fig:RC_IN1K}, we show the scatter plots of the 500 architectures' supervised classification accuracy (horizontal axis) and pretext task accuracy (vertical axis) on CIFAR-10 and ImageNet, respectively. We see that this rank correlation is typically higher than 0.8, regardless of the dataset, the search space, and the pretext task. This type of consistency and robustness indicates that this phenomenon is general, as opposed to dataset/search space specific. 

The same experiment is performed on both the DARTS and the NAS-Bench-101 search spaces. The rank correlations on the NAS-Bench-101 search space are generally higher than those on the DARTS search space. A possible explanation is that the architectures in NAS-Bench-101 are more diverse, and consequently their accuracies have larger gaps, \ie, are less affected by training noise. 

Interestingly, we observe that among the three pretext tasks, colorization consistently has the lowest correlation on CIFAR-10, but the highest correlation on ImageNet. We suspect this is because the small images in CIFAR-10 make the learning of per-pixel colorization difficult, and consequently the performance after training is noisy.

\subsubsection{High rank correlation between supervised accuracy and pretext accuracy \underline{across} datasets.}
\label{sec:sample-based-across}

In Figure~\ref{fig:RC_across} we show the \emph{across dataset} rank correlation analysis, where the pretext task accuracy is measured on CIFAR-10, but the supervised classification accuracy is measured on ImageNet. 

On the DARTS search space (top panel), despite the image distribution shift brought by different datasets, for each of the three pretext tasks (left three plots), the correlation remains consistently high ($\app$0.8). This shows that across the entire search space, the relative ranking of an architecture is likely to be similar, whether under the unsupervised pretext task accuracy or under the supervised target task accuracy. 
In the rightmost panel, we compare with a scenario where instead of using an unsupervised pretext task, we use a proxy task of CIFAR-10 supervised classification. The correlation is $\rho$ $=$ 0.90 in this case. As the CIFAR-10 supervised classification is a commonly used proxy task in existing NAS literature \cite{zoph2018learning}, it gives a reference on $\rho$'s value.

In the bottom panel we show more analysis of this kind by replacing the search space with NAS-Bench-101. Although the architectures are quite different, the observations are similar. In all cases, the self-supervised pretext task accuracy is highly correlated to the supervised classification accuracy.

\subsubsection{Better pretext accuracy translates to better supervised accuracy.}

\begin{figure}[t]\centering
\begin{subfigure}[t]{0.4\textwidth}
\captionsetup{justification=centering}
\centering
\includegraphics[width=1.0\linewidth]{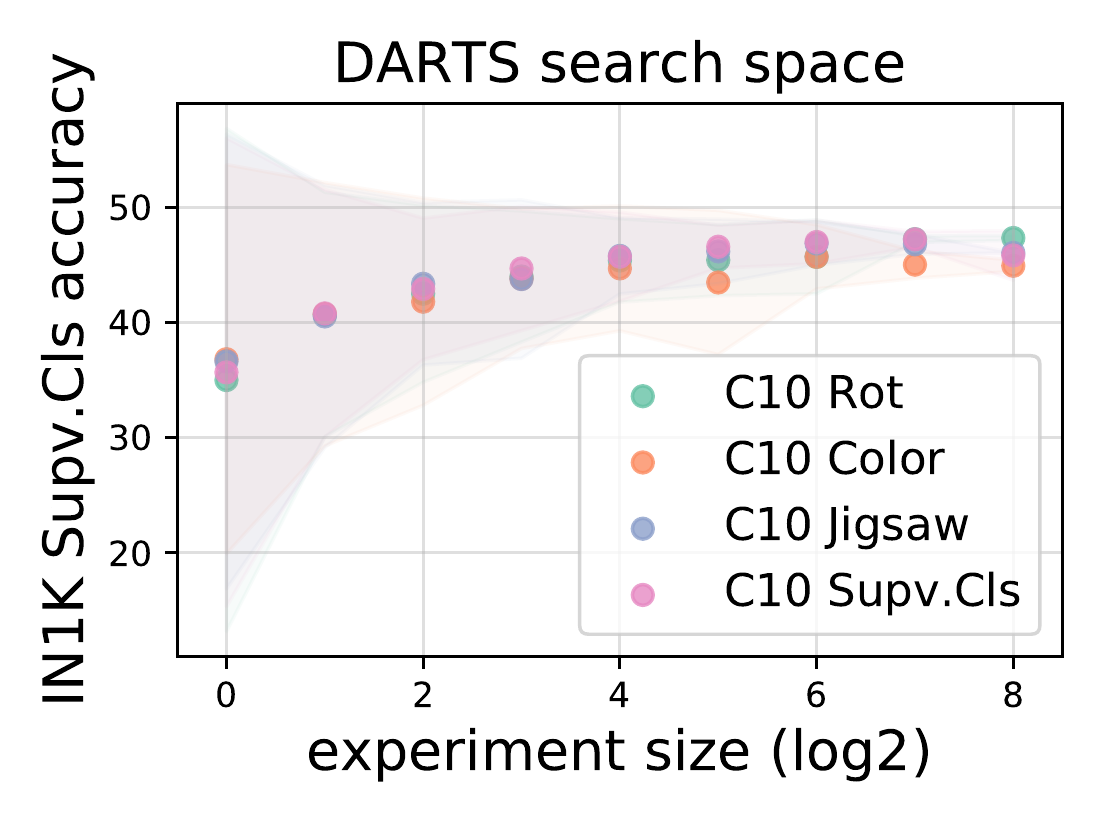}
\end{subfigure}
~
\begin{subfigure}[t]{0.4\textwidth}
\captionsetup{justification=centering}
\centering
\includegraphics[width=1.0\linewidth]{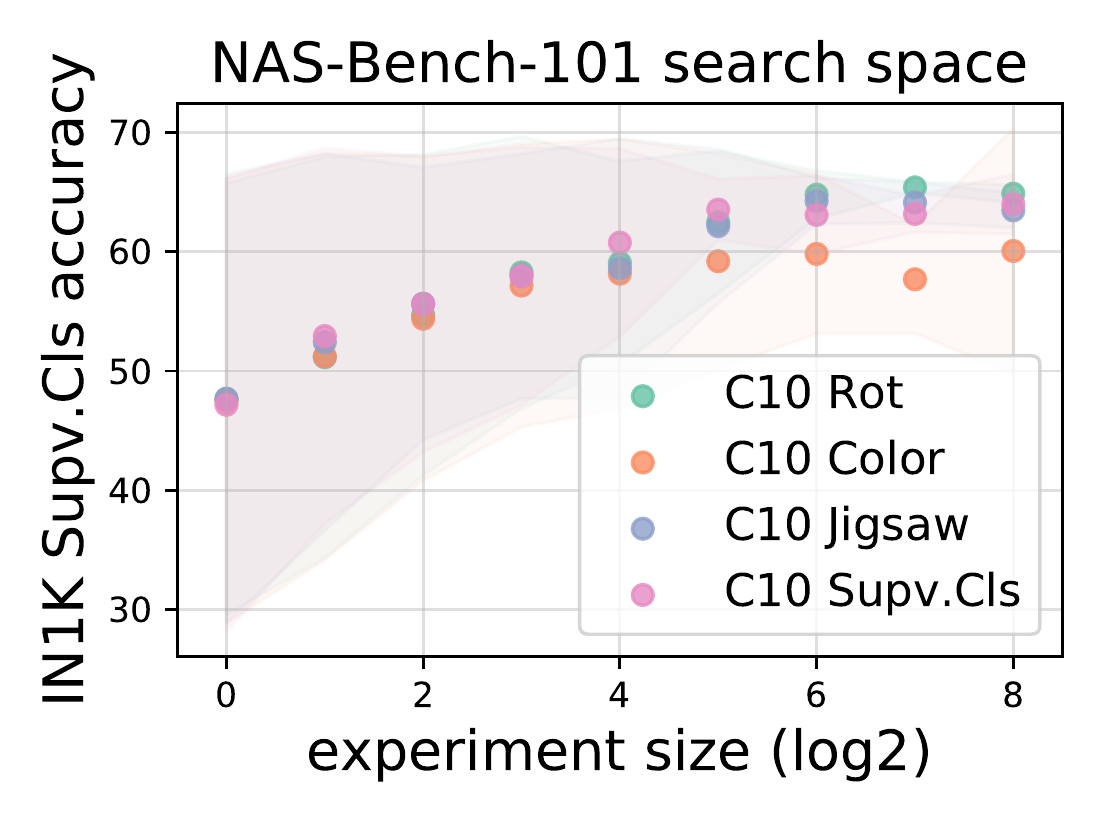}
\end{subfigure}
\caption{\textbf{Random experiment efficiency curves}. Left panel: DARTS search space. Right panel: NAS-Bench-101 search space. We show the range of ImageNet classification accuracies of top architectures identified by the three pretext tasks and the supervised task under various experiment sizes. See text for more details. }
\label{fig:bench_random_search}
\end{figure}

In addition to the rank correlation analysis, we also perform the random experiment analysis. Figure~\ref{fig:bench_random_search} shows the \emph{random experiment efficiency} curve for DARTS and NAS-Bench-101 search spaces. 
Again, the pretext accuracies are obtained on CIFAR-10, and the target accuracies are from ImageNet.
By design of this experiment, as the experiment size $m$ increases, the \emph{pretext} accuracies of the $\lceil n/m \rceil$ architectures should increase. Figure~\ref{fig:bench_random_search} shows that the \emph{target} accuracies of these $\lceil n/m \rceil$ architectures also increase with $m$.
In addition, at each experiment size, most unsupervised pretext objectives perform similarly compared to the commonly used supervised CIFAR-10 proxy. The overall trends are also comparable. 
This shows that the architecture rankings produced with and without labels are not only correlated across the entire search space, but also towards the top of the search space, which is closer to the goal of \unnas{}. 

\section{Search-Based Experiments}
\label{sec:search-based}

\subsection{Experimental Design}

In \emph{search-based experiments}, the idea is to run a well-established NAS algorithm, except that we make the minimal modification of replacing its supervised search objective with an unsupervised one. Following the \unnas{} setup, we then examine (by training from scratch) how well these unsupervisedly discovered architectures perform on supervised target tasks. Since all other variables are controlled to be the same, search-based experiments can easily compare between the supervised and unsupervised counterparts of a NAS algorithm, which can help reveal the importance of labels. 

The NAS algorithm we adopt is DARTS (short for ``differentiable architecture search")~\cite{liu2018darts} for its simplicity. DARTS formulates the activation tensor selection and operation selection as a categorical choice, implemented as a softmax function on a set of continuous parameters (named $\alpha$). These parameters are trained in a similar fashion as the architecture weights, by backpropagation from a loss function. After this training, the softmax outputs are discretized and produce an architecture.

The self-supervised objectives that we consider are, still, those described in Section~\ref{sec:pretext}: rotation prediction (\rot), colorization (\col), and solving jigsaw puzzles (\jig). For comparison, we also perform NAS search with supervised objectives, \eg classification (\cls) or semantic segmentation (\seg). To help distinguish, we name the method \nasdarts{} if the search objective is supervised, and \unnasdarts{} if the search objective is unsupervised. 

\subsection{Implementation Details}

\paragraph{Search phase.} 
We use three different datasets for architecture search: ImageNet-1K (IN1K)~\cite{Deng2009}, ImageNet-22K (IN22K)~\cite{Deng2009}, and Cityscapes~\cite{Cordts2016}. IN1K is the standard ImageNet benchmark dataset with 1.2M images from 1K categories. IN22K is the full ImageNet dataset that has $\app$14M images from 22K categories. Cityscapes is a dataset of street scenes that has drastically different image statistics. 
Note that \unnasdarts{} will only access the images provided in the dataset, while \nasdarts{} will additionally access the (semantic) labels provided in the respective dataset. The search phase will operate only within the training split, without accessing the true validation or test split. 

We report in Appendix the hyper-parameters we used. One major difference between our experiments and DARTS~\cite{liu2018darts} is that the images in the search datasets that we consider are much larger in size. We use 224$\times$224 random crops for search on IN1K/IN22K, and 312$\times$312 for search on Cityscapes following~\cite{liu2019auto}. To enable DARTS training with large input images, we use 3 stride-2 convolution layers at the beginning of the network to reduce spatial resolution. This design, together with appropriately chosen number of search epochs (see Appendix), allows \unnas{} search to be efficient ($\app$2 GPU days on IN1K/Cityscapes, $\app$10 GPU days on IN22K, regardless of task) despite running on larger images. 

\paragraph{Evaluation phase.}
We use two distinct datasets and tasks for UnNAS evaluation: (1) ImageNet-1K (IN1K) for image classification. The performance metric is top-1 accuracy on the IN1K validation set. (2) Cityscapes for semantic segmentation. We use the \texttt{train\_fine} set (2975 images) for training. The performance metric is mean Intersection-over-Union (mIoU) evaluated on the \texttt{val} set (500 images). 

For IN1K evaluation, we fix depth to 14 and adjust the width to have \#FLOPs $\in$ [500,600]M. Models are trained for 250 epochs with an auxiliary loss weighted by 0.4, batch size 1024 across 8 GPUs, cosine learning rate schedule \cite{loshchilov2016sgdr} with initial value 0.5, and 5 epochs of warmup. 
For Cityscapes evaluation, we fix depth to 12 and adjust the width to have \#Params $\in$ [9.5,10.5]M. We train the network for 2700 epochs, with batch size 64 across 8 GPUs, cosine learning rate schedule with initial value 0.1. 
For both ImageNet and Cityscapes evaluations, we report the mean and standard deviation of 3 independent trainings of the same architecture.
More implementation details are described in Appendix. 

We note that under our definition of the UnNAS setup, the same dataset should not be used for both search and evaluation (because this scenario is unrealistic); We provide the IN1K$\rightarrow$IN1K and Cityscapes$\rightarrow$Cityscapes results purely as a reference. Those settings are analogous to the linear classifier probe for IN1K in conventional unsupervised learning research.

\subsection{Results}

In search-based experiments, the architectures are evaluated on both ImageNet classification, summarized in Table~\ref{tab:imagenet250}, and Cityscapes semantic segmentation, summarized in Table~\ref{tab:cityscapes}. 
We provide visualization of all \nasdarts{} and \unnasdarts{} cell architectures in Appendix.

\newcolumntype{H}{>{\setbox0=\hbox\bgroup}c<{\egroup}@{}}
\begin{table}[t]\centering
\tablestyle{4.5pt}{1.05}\begin{tabular}{ll|lH|cc}
method & search dataset \& task & top-1 acc. & top-5 acc. & \ft{7}FLOPs (M) & \ft{7}params (M) \\
\shline
NAS-\texttt{DARTS}~\cite{liu2018darts} & CIFAR-10 \cls{} & 73.3 & 91.3 & 574 & 4.7 \\
NAS-\texttt{P-DARTS}~\cite{chen2019progressive} & CIFAR-10 \cls{} & 75.6 & 92.6 & 557 & 4.9 \\
NAS-\texttt{PC-DARTS}~\cite{xu2019pc} & CIFAR-10 \cls{} & 74.9 & 92.2 & 586 & 5.3 \\
NAS-\texttt{PC-DARTS}~\cite{xu2019pc} & IN1K \cls{} & 75.8 & 92.7 & 597 & 5.3 \\
NAS-\texttt{DARTS}$^\dagger$ & CIFAR-10 \cls{} & 74.9\pmstd{0.08} & 92.2\pmstd{0.06} & 538 & 4.7 \\
\hline 
\hline
{NAS-\texttt{DARTS}} & IN1K \cls{} & {76.3}\pmstd{0.06} & {92.9}\pmstd{0.04} & 590 & 5.3 \\
\grayfy{{\unnas-\texttt{DARTS}}} & \grayfy{IN1K \rot{}} & \grayfy{{75.8}\pmstd{0.18}} & \grayfy{{92.7}\pmstd{0.16}} & \grayfy{558} & \grayfy{5.1} \\
\grayfy{{\unnas-\texttt{DARTS}}} & \grayfy{IN1K \col{}} & \grayfy{{75.7}\pmstd{0.12}} & \grayfy{{92.6}\pmstd{0.16}} & \grayfy{547} & \grayfy{4.9} \\
\grayfy{{\unnas-\texttt{DARTS}}} & \grayfy{IN1K \jig{}} & \grayfy{{75.9}\pmstd{0.15}} & \grayfy{{92.8}\pmstd{0.10}} & \grayfy{567} & \grayfy{5.2} \\  
\hline
{NAS-\texttt{DARTS}} & IN22K \cls{} & {75.9}\pmstd{0.09} & {92.7}\pmstd{0.08} & 585 & 5.2 \\
{\unnas-\texttt{DARTS}} & IN22K \rot{} & {75.7}\pmstd{0.23} & {92.7}\pmstd{0.11} & 549 & 5.0 \\
{\unnas-\texttt{DARTS}} & IN22K \col{} & {75.9}\pmstd{0.21} & {92.8}\pmstd{0.10} & 547 & 5.0 \\
{\unnas-\texttt{DARTS}} & IN22K \jig{} & {75.9}\pmstd{0.31} & {92.8}\pmstd{0.13} & 559 & 5.1 \\  
\hline
{\nasdarts} & Cityscapes \seg{} & {75.8}\pmstd{0.13} & {92.6}\pmstd{0.17} & 566 & 5.1 \\
{\unnasdarts} & Cityscapes \rot{} & {75.9}\pmstd{0.19} & {92.7}\pmstd{0.07} & 554 & 5.1 \\
{\unnas-\texttt{DARTS}} & Cityscapes \col{} & {75.2}\pmstd{0.15} & {92.4}\pmstd{0.05} & 594 & 5.1 \\
{\unnas-\texttt{DARTS}} & Cityscapes \jig{} & {75.5}\pmstd{0.06} & {92.6}\pmstd{0.08} & 566 & 5.0 \\  
\end{tabular}
  \caption{\textbf{ImageNet-1K classification results of the architectures searched by NAS and \unnas{} algorithms}. \grayfy{Rows in gray} correspond to invalid \unnas{} configurations where the search and evaluation datasets are the same. $\dagger$ is our training result of the DARTS architecture released in \cite{liu2018darts}.}
\label{tab:imagenet250}
\end{table}

\begin{table}[t]\centering
\tablestyle{4.5pt}{1.05}\begin{tabular}{ll|l|cc}
method & search dataset \& task & mIoU & \ft{7}FLOPs (B) & \ft{7}params (M) \\
\shline
\nasdarts $^\dagger$ & CIFAR-10 \cls{} & 72.6\pmstd{0.55} & 121 & 9.6 \\
\hline
\hline
{\nasdarts} & IN1K \cls{}  & {73.6}\pmstd{0.31} & 127 & 10.2 \\
{\unnasdarts} & IN1K \rot{} & {73.6}\pmstd{0.29} & 129 & 10.4 \\
{\unnasdarts} & IN1K \col{} & {72.2}\pmstd{0.56} & 122 & 9.7 \\
{\unnasdarts} & IN1K \jig{} & {73.1}\pmstd{0.17}  & 129 & 10.4 \\  
\hline
{\nasdarts} & IN22K \cls{} & {72.4}\pmstd{0.29} & 126 & 10.1 \\
{\unnasdarts} & IN22K \rot{} & {72.9}\pmstd{0.23} & 128 & 10.3 \\
{\unnasdarts} & IN22K \col{} & {73.6}\pmstd{0.41} & 128 & 10.3 \\
{\unnasdarts} & IN22K \jig{} & {73.1}\pmstd{0.59} & 129 & 10.4 \\  
\hline
{\nasdarts} & Cityscapes \seg{} & {72.4}\pmstd{0.15} & 128 & 10.3 \\
\grayfy{{\unnasdarts}} & \grayfy{Cityscapes \rot{}} & \grayfy{{73.0}\pmstd{0.25}}  & \grayfy{128} & \grayfy{10.3} \\
\grayfy{{\unnasdarts}} & \grayfy{Cityscapes \col{}} & \grayfy{{72.5}\pmstd{0.31}} & \grayfy{122} & \grayfy{9.5} \\
\grayfy{\textbf{{\unnasdarts}}} & \grayfy{\textbf{Cityscapes \jig{}}} & \grayfy{\textbf{{74.1}\pmstd{0.39}}} & \grayfy{\textbf{128}} & \grayfy{\textbf{10.2}} \\  
\end{tabular}
\caption{\textbf{Cityscapes semantic segmentation results of the architectures searched by NAS and \unnas{} algorithms}. These are trained from scratch: there is no fine-tuning from ImageNet checkpoint. \grayfy{Rows in gray} correspond to an illegitimate setup where the search dataset is the same as the evaluation dataset. $\dagger$ is our training result of the DARTS architecture released in \cite{liu2018darts}. }
\label{tab:cityscapes}
\end{table}

\subsubsection{\unnas{} architectures perform competitively to supervised counterparts.}

We begin by comparing \nasdarts{} and \unnasdarts{} when they are performed on the same search dataset. This would correspond to every four consecutive rows in Table~\ref{tab:imagenet250} and Table~\ref{tab:cityscapes}, grouped together by horizontal lines.

As discussed earlier, strictly speaking the IN1K$\rightarrow$IN1K experiment is not valid under our definition of \unnas{}. For reference, we \grayfy{gray out} these results in Table~\ref{tab:imagenet250}. \nasdarts{} on IN1K dataset has the highest performance among our experiments, achieving a top-1 accuracy of $76.3\%$. However, the \unnas{} algorithm variants with \rot{}, \col{}, \jig{} objectives all perform very well (achieving $75.8\%$, $75.7\%$ and $75.9\%$ top-1 accuracy, respectively), closely approaching the results obtained by the supervised counterpart. This suggests it might be desirable to perform architecture search on the target dataset directly, as also observed in other work~\cite{cai2018proxylessnas}.

Two valid \unnas{} settings include IN22K$\rightarrow$IN1K and Cityscapes$\rightarrow$IN1K for architecture search and evaluation. For IN22K$\rightarrow$IN1K experiments, NAS and \unnas{} results across the board are comparable. For Cityscapes$\rightarrow$IN1K experiments, among the \unnas{} architectures, \rot{} and \jig{}  perform well, once again achieving results comparable to the supervised search. However, there is a drop for \unnasdarts{} search with \col{} objective (with a $75.2\%$ top-1 accuracy). We hypothesize that this might be owing to the fact that the color distribution in Cityscapes images is not as diverse: the majority of the pixels are from \emph{road} and \emph{ground} categories of gray colors.

In general, the variances are higher for Cityscapes semantic segmentation (Table~\ref{tab:cityscapes}), but overall \unnasdarts{} architectures still perform competitively to \nasdarts{} architectures, measured by mIoU. For the Cityscapes$\rightarrow$Cityscapes experiment, we observe that searching with segmentation objective directly leads to inferior result (mean $72.4\%$ mIoU), compared to the architectures searched for ImageNet classification tasks. This is different from what has been observed in Table~\ref{tab:imagenet250}. However, under this setting, our \unnas{} algorithm, in particular the one with the \jig{} objective shows very promising results (mean $74.1\%$ mIoU). In fact, when the search dataset is IN22K or Cityscapes, all \unnasdarts{} architectures perform \emph{better} than the \nasdarts{} architecture. This is the opposite of what was observed in IN1K$\rightarrow$IN1K. Results for Cityscapes$\rightarrow$Cityscapes are \grayfy{grayed out} for the same reason as before (invalid under \unnas{} definition).

\subsubsection{NAS and \unnas{} results are robust across a large variety of datasets and tasks.}

The three search datasets that we consider are of different nature. For example, IN22K is 10 times larger than IN1K, and Cityscapes images have a markedly different distribution than those in ImageNet.
In our experiments, \nasdarts{}/\unnasdarts{} architectures searched on IN22K do not significantly outperform those searched on IN1K, meaning that they do not seem to be able to enjoy the benefit of having more abundant images. This reveals new opportunities in designing better algorithms to exploit bigger datasets for neural architecture search. For Cityscapes$\rightarrow$IN1K experiments, it is interesting to see that after switching to a dataset with markedly distinct search images (urban street scenes), we are still able to observe decent performance. The same goes for the reverse direction IN1K/IN22K$\rightarrow$Cityscapes, which implies that the search does not severely overfit to the images from the dataset. 

In addition to this robustness to the \emph{search dataset} distribution, NAS and \unnas{} also exhibit robustness to \emph{target dataset and task}. Classification on ImageNet and segmentation on Cityscapes are different in many ways, but among different combinations of search dataset and task (whether supervised or unsupervised), we do not observe a case where the same architecture performs well on one but poorly on the other.

\subsubsection{\unnas{} outperforms previous methods.}

Finally, we compare our \unnasdarts{} results against existing works. 
We first note that we are able to achieve a better baseline number with the \nasdarts{} architecture (searched with the CIFAR-10 proxy) compared to what was reported in~\cite{liu2018darts}. This is mainly due to better hyper-parameter tuning we adopt from~\cite{chen2019progressive} for the \emph{evaluation phase} model training. This baseline sets up a fair ground for all the \unnas{} experiments; we use the same evaluation phase hyper-parameters across different settings. 

On ImageNet, our \unnasdarts{} architectures can comfortably outperform this baseline by up to $1\%$ classification accuracy. In fact, the extremely competitive \unnasdarts{} results also outperform the previous best result ($75.8\%$) on this search space, achieved with a more sophisticated NAS algorithm \cite{xu2019pc}.

On Cityscapes, there have not been many works that use DARTS variants as the backbone. The closest is Auto-DeepLab \cite{liu2019auto}, but we use a lighter architecture (in that we do not have the Decoder) and shorter training iterations, so the results are not directly comparable. Nonetheless, according to our evaluation, the \unnas{} architectures perform favorably against the DARTS architecture released in \cite{liu2018darts} (discovered with the CIFAR-10 proxy). The best \unnasdarts{} variant (Cityscapes \jig{}) achieves $74.1\%$ mIoU, which outperforms this baseline by $1.5\%$ on average. 
Overall, our experiments demonstrate that exploring neural architecture search with unsupervised/self-supervised objectives to improve target task performance might be a fruitful direction. 

Outperforming previous methods was far from the original goal of our study. Nonetheless, the promising results of \unnas{} suggest that in addition to developing new \emph{algorithms} and finding new \emph{tasks}, the role of \emph{data} (in our case, more/larger images) and \emph{paradigm} (in our case, no human annotations) is also worth attention in future work on neural architecture search.

\section{Discussion}

In this paper, we challenge the common practice in neural architecture search and ask the question: do we really need labels to successfully perform NAS? We approach this question with two sets of experiments. In sample-based experiments, we discover the phenomenon that the architecture rankings produced with and without labels are highly correlated. In search-based experiments, we show that the architectures learned without accessing labels perform competitively, not only relative to their supervised counterpart, but also in terms of absolute performance. In both experiments, the observations are consistent and robust across various datasets, tasks, and/or search spaces. Overall, the findings in this paper indicate that labels are \emph{not} necessary for neural architecture search. 

How to learn and transfer useful representations to subsequent tasks in an unsupervised fashion has been a research topic of extensive interest, but the discovery of neural network architectures has been driven solely by supervised tasks. As a result, current NAS products or AutoML APIs typically have the strict prerequisite for users to ``\emph{put together a training dataset of labeled images}''~\cite{AutoML_API}. An immediate implication of our study is that the job of the user could potentially be made easier by dropping the labeling effort. In this sense, \unnas{} could be especially beneficial to the many applications where data constantly comes in at large volume but labeling is costly.

At the same time, we should still ask: if not labels, then what factors are needed to reveal a good architecture? A meaningful unsupervised task seems to be important, though the several pretext tasks considered in our paper do not exhibit significant difference in either of the two experiments. In the future we plan to investigate even more and even simpler unsupervised tasks. Another possibility is that the architecture quality is mainly decided by the image statistics, and since the datasets that we consider are all natural images, the correlations are high and the results are comparable. This hypothesis would also suggest an interesting, alternative direction: that instead of performing NAS again and again for every specific labeled task, it may be more sensible to perform NAS once on large amounts of unlabeled images that capture the image distribution.


%
%
\bibliographystyle{splncs04}
\bibliography{unnas}

\clearpage
\appendix

\section{Additional Details for Sample-Based Experiments}

In sampling the 500 DARTS architectures, we only keep those whose \#Params and \#FLOPs both fall into [80\%, 120\%] that of ResNet-56. Doing so prevents the influence of model capacity on the rank correlations, but the downside is that these architectures may not be representative of the entire search space. Therefore, in sampling the 500 NAS-Bench-101 architectures, we used a complementary strategy where we first rank all architectures based on the released accuracy, and then sample 50 architectures from each 10th percentile.

\section{Additional Details for Search-Based Experiments}

\subsection{Search Phase}

Recall that in the search phase of \nasdarts{}/\unnasdarts{}, we consider 3 datasets: ImageNet-1K (IN1K), ImageNet-22K (IN22K), Cityscapes. 

For IN1K, we follow \cite{liu2019auto,chen2019progressive} to postpone updating architecture parameters $\alpha$ for half of the total search epochs.
For IN1K/IN22K, the search phase lasts 2 epochs for IN1K and 1 epoch for IN22K. Since IN22K is approximately 10 times larger than IN1K ($\app$14M vs $\app$1.2M), the search on IN22K is approximately 5 times longer than the search on IN1K. Batch size is 64, learning rate is 0.1 (cosine schedule), weight decay is 0.00003. 
For Cityscapes, the search phase lasts 400 epochs. Batch size is 32, learning rate is 0.1 (cosine schedule), weight decay is 0.0003. 
Other than those listed here and in the main paper, all hyperparameters follow those used in the original DARTS. 
In a few settings where the batch size above will exceed 32GB GPU memory, we divide batch size and learning rate (both for weights and for architecture parameters $\alpha$) by 2. 

\subsection{Evaluation Phase}

When evaluating an architecture on Cityscapes semantic segmentation, since the task is pixel-level classification instead of image-level classification, we need to make minimal but necessary modifications. Different from the network used in IN1K classification, we (1) replace the last stride 2 layer with stride 1, to increase spatial resolution; and (2) remove the global average pooling and replace the fully connected classifier with the Atrous Spatial Pyramid Pooling (ASPP) module~\cite{chen2017rethinking} followed by bilinear upsampling to produce per-pixel classification at the original resolution. These are the same modifications the segmentation framework DeepLabv3~\cite{chen2017rethinking} made to a ResNet backbone. 

\section{\nasdarts{} and \unnasdarts{} Architectures}

Here we visualize all the \nasdarts{} and \unnasdarts{} cell architectures: searched on ImageNet-1K (Figure~\ref{fig:darts_in1k_archs}), ImageNet-22K (Figure~\ref{fig:darts_in22k_archs}), Cityscapes (Figure~\ref{fig:darts_cityscapes_archs}). 

\newcommand{\figw}{0.9\linewidth}

\begin{figure}[!htp]\centering

\begin{subfigure}[t]{0.24\textwidth}
\centering
\includegraphics[width=\figw]{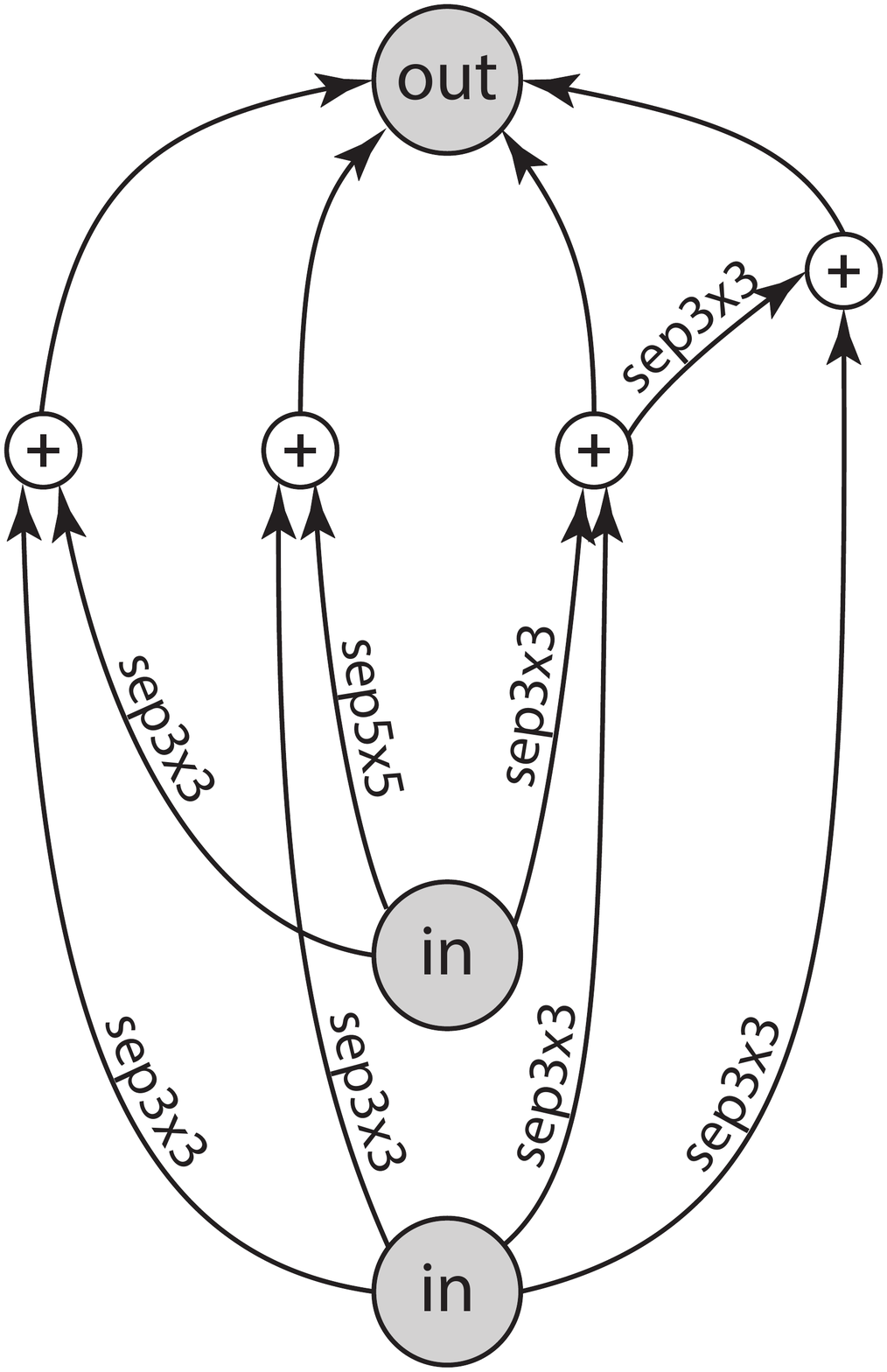}
\includegraphics[width=\figw]{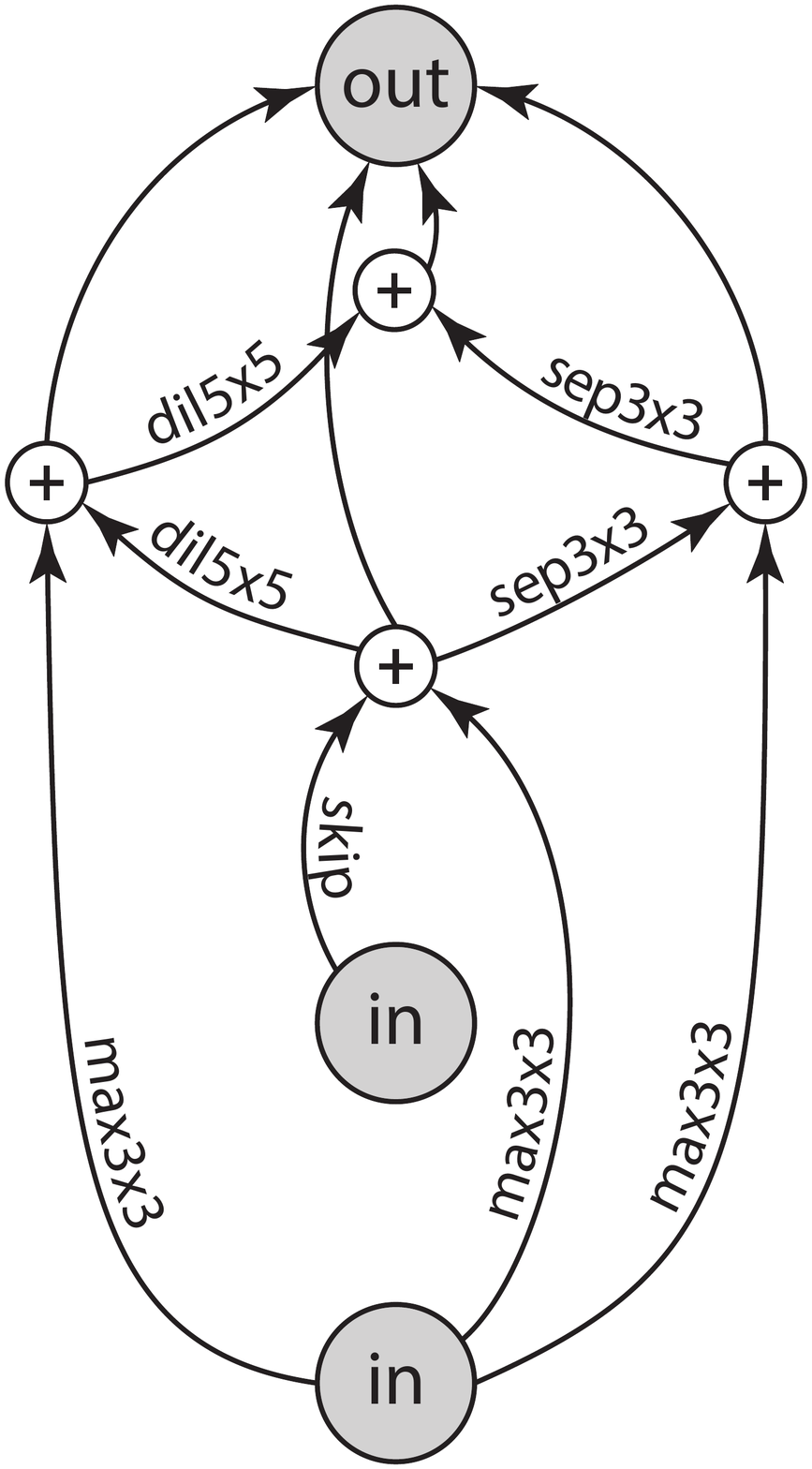}
\caption{NAS \cls{}}
\end{subfigure}
\begin{subfigure}[t]{0.24\textwidth}
\centering
\includegraphics[width=\figw]{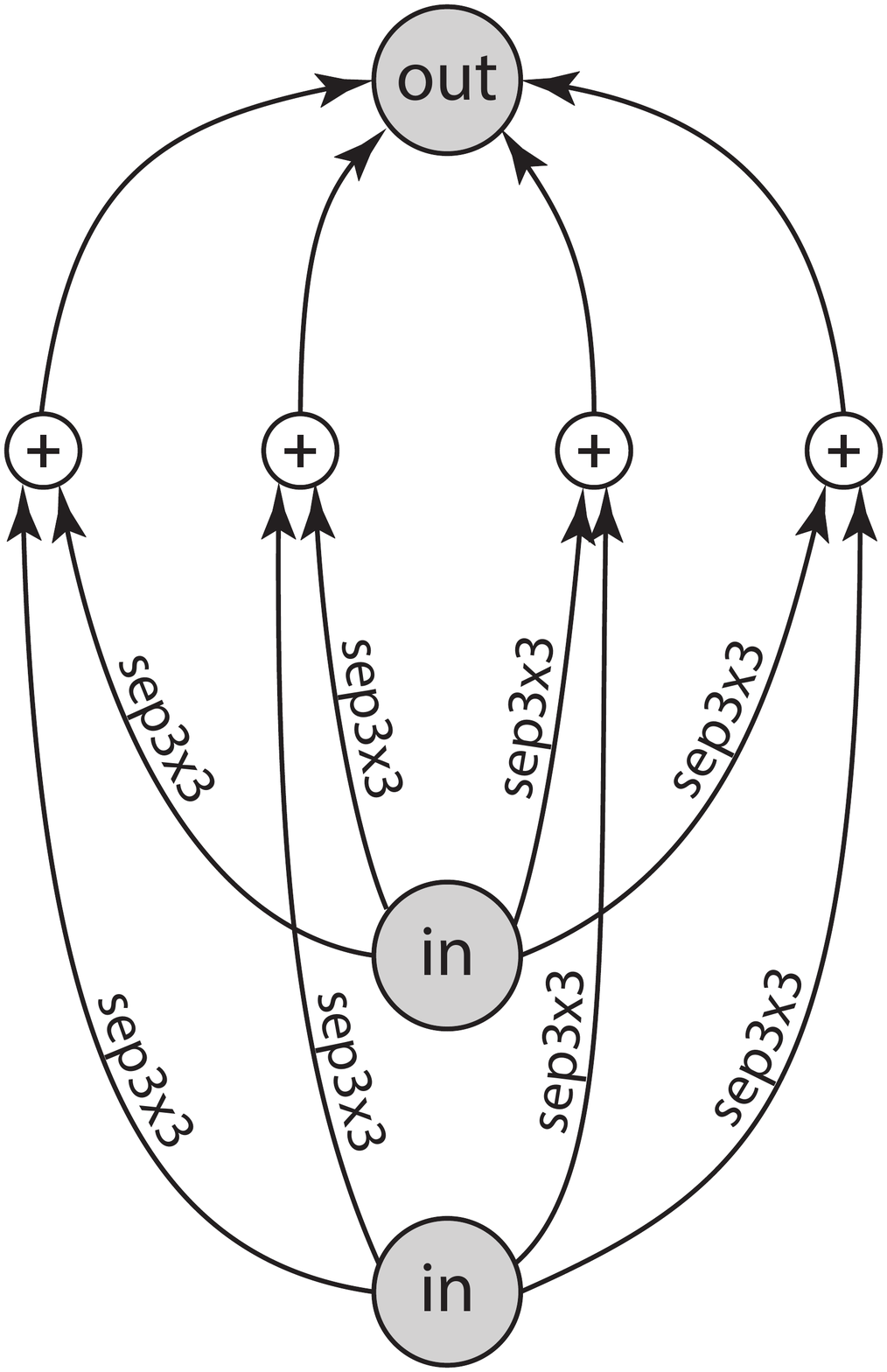}
\includegraphics[width=\figw]{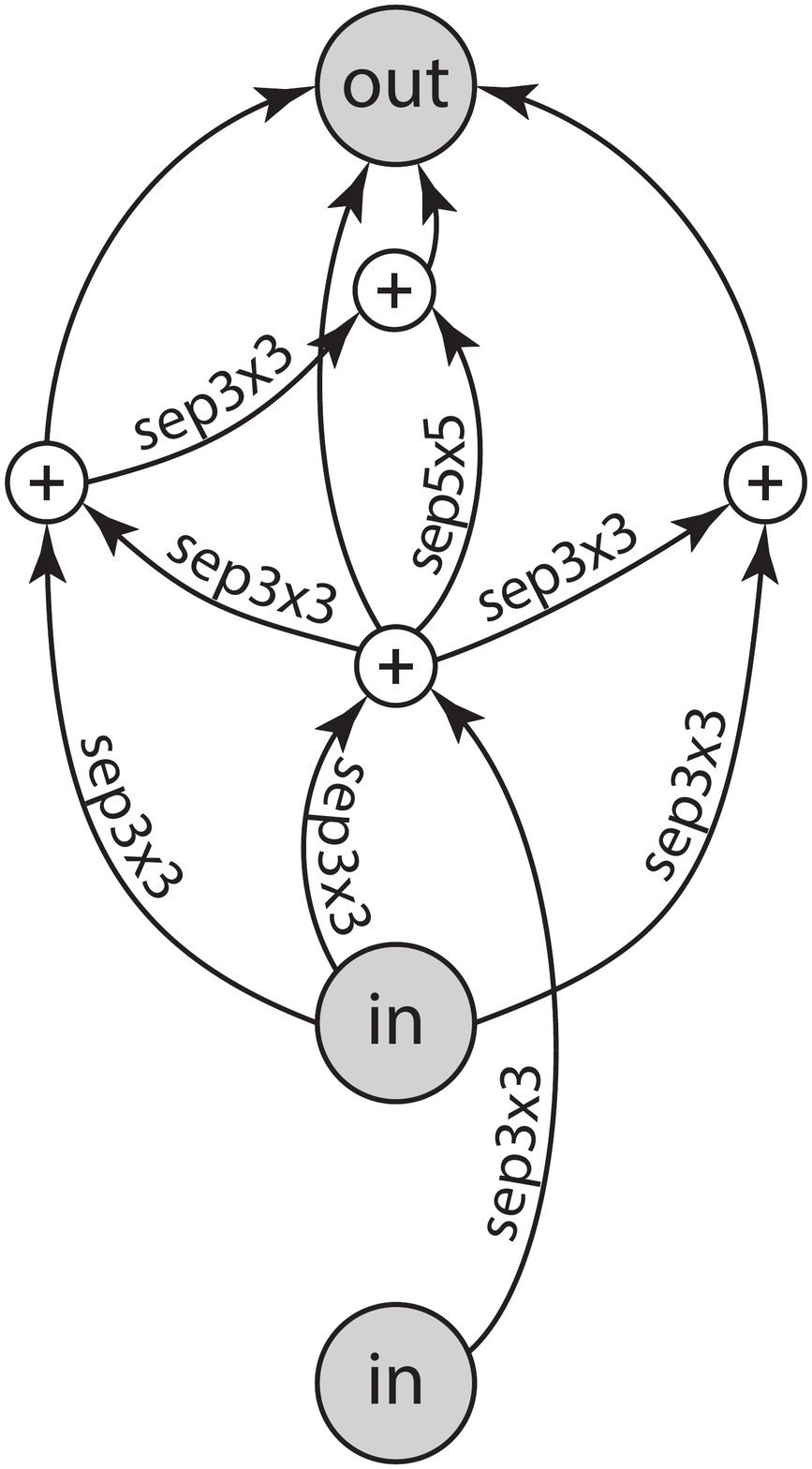}
\caption{\unnas{} \rot{}}
\end{subfigure}
\begin{subfigure}[t]{0.24\textwidth}
\centering
\includegraphics[width=\figw]{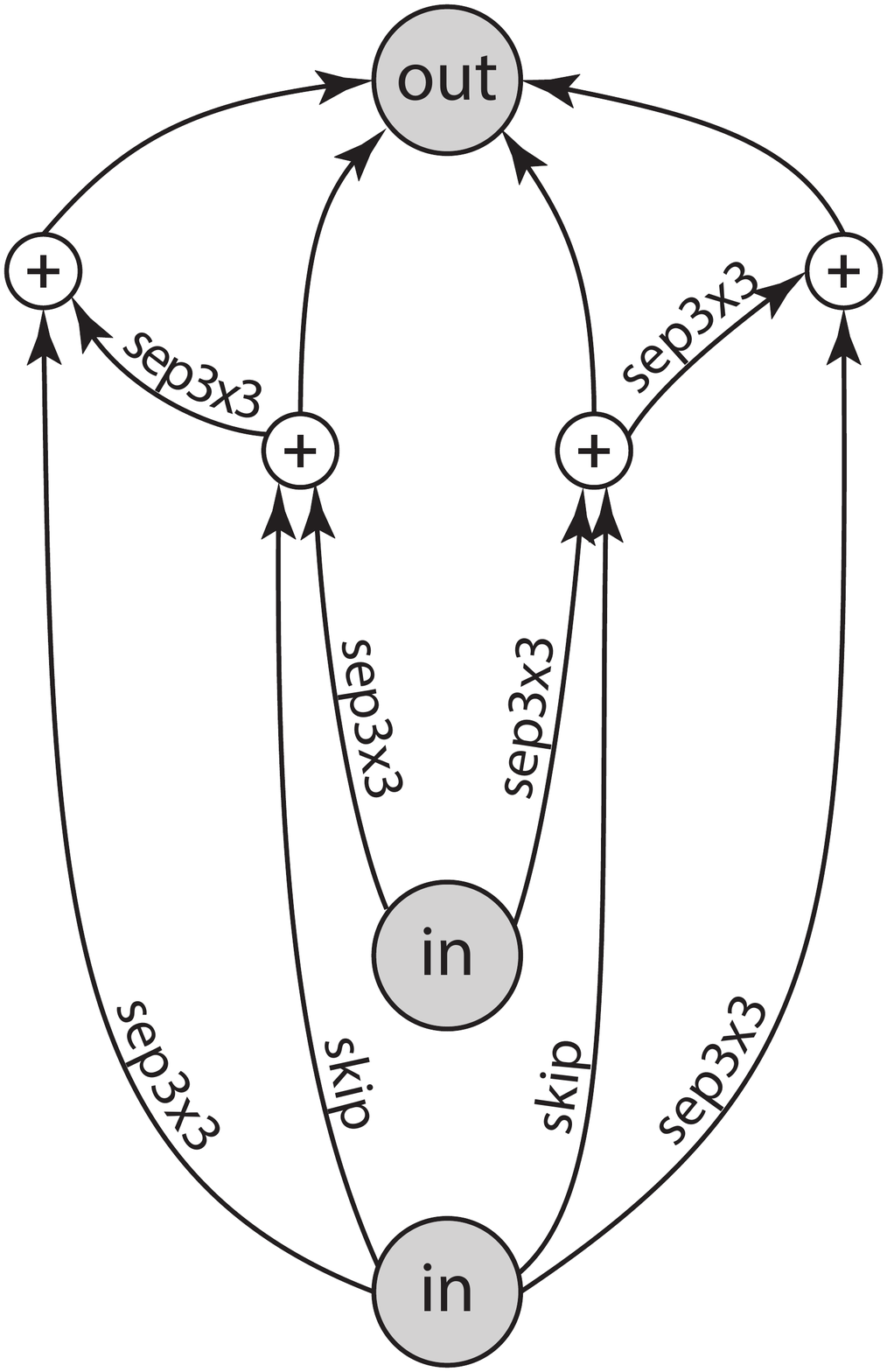}
\includegraphics[width=\figw]{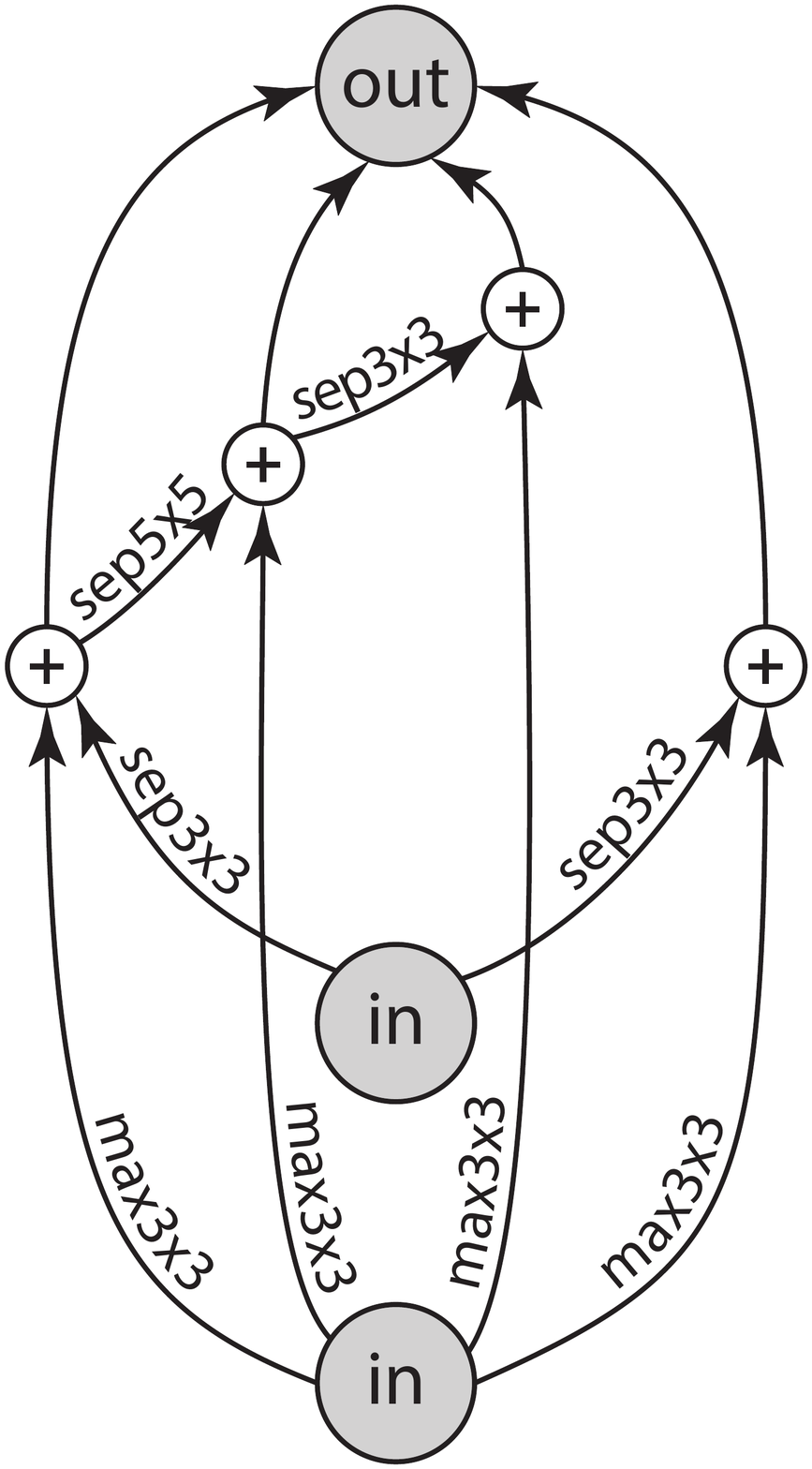}
\caption{\unnas{} \col{}}
\end{subfigure}
\begin{subfigure}[t]{0.24\textwidth}
\centering
\includegraphics[width=\figw]{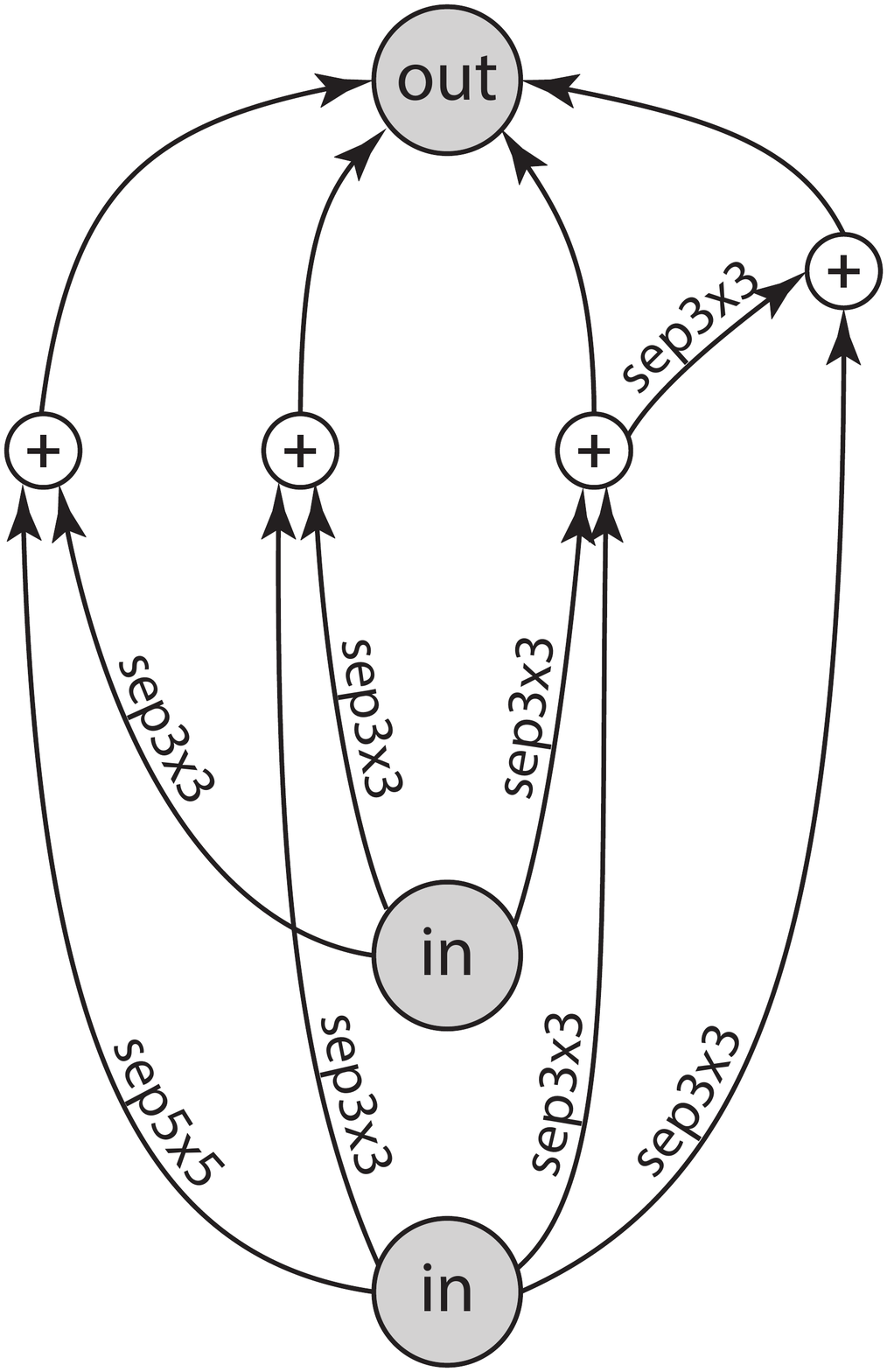}
\includegraphics[width=\figw]{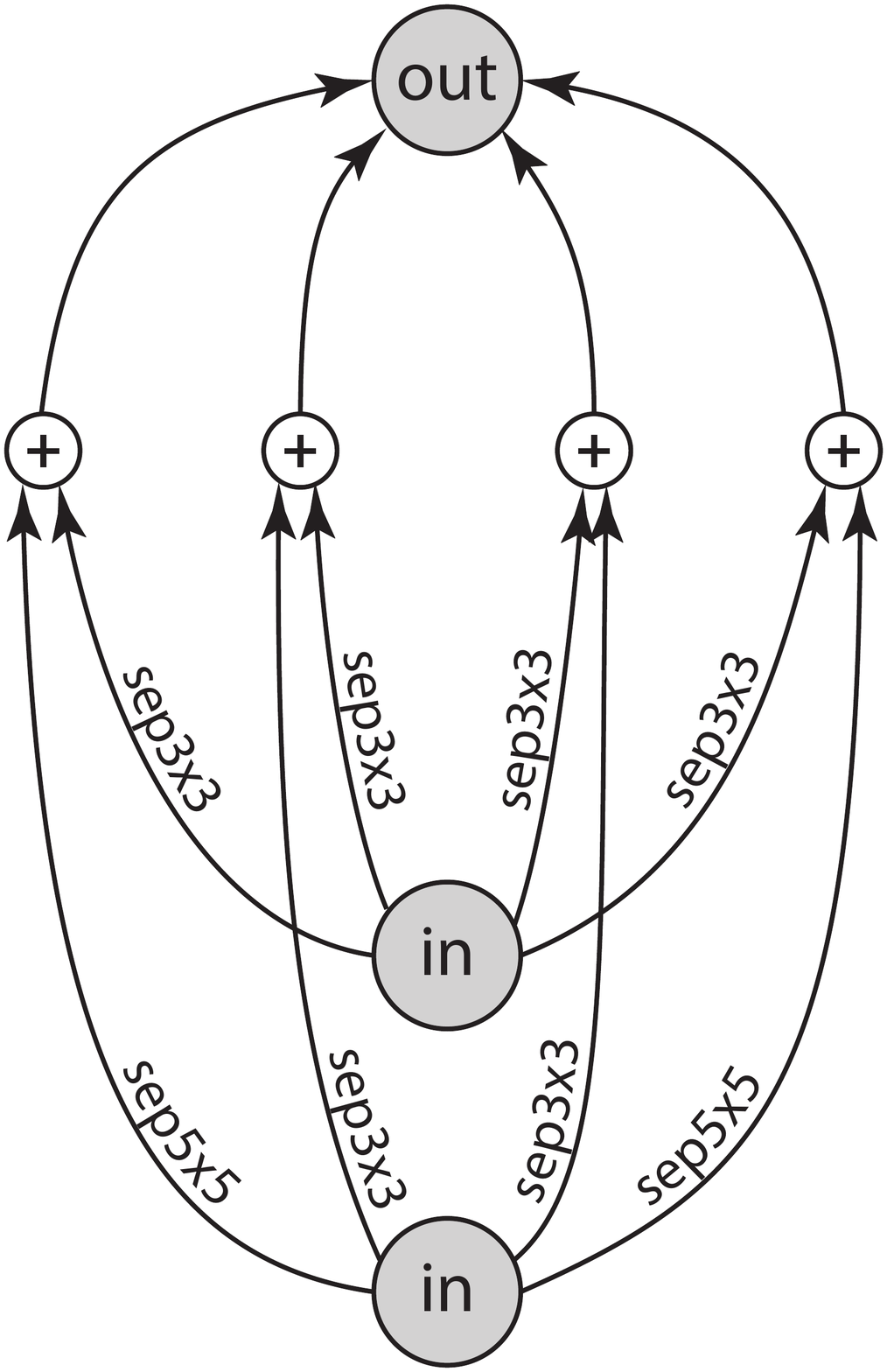}
\caption{\unnas{} \jig{}}
\end{subfigure}

\caption{Cell architectures (normal and reduce) searched on ImageNet-1K. }
\label{fig:darts_in1k_archs}
\end{figure}

\begin{figure}[!htp]\centering

\begin{subfigure}[t]{0.24\textwidth}
\centering
\includegraphics[width=\figw]{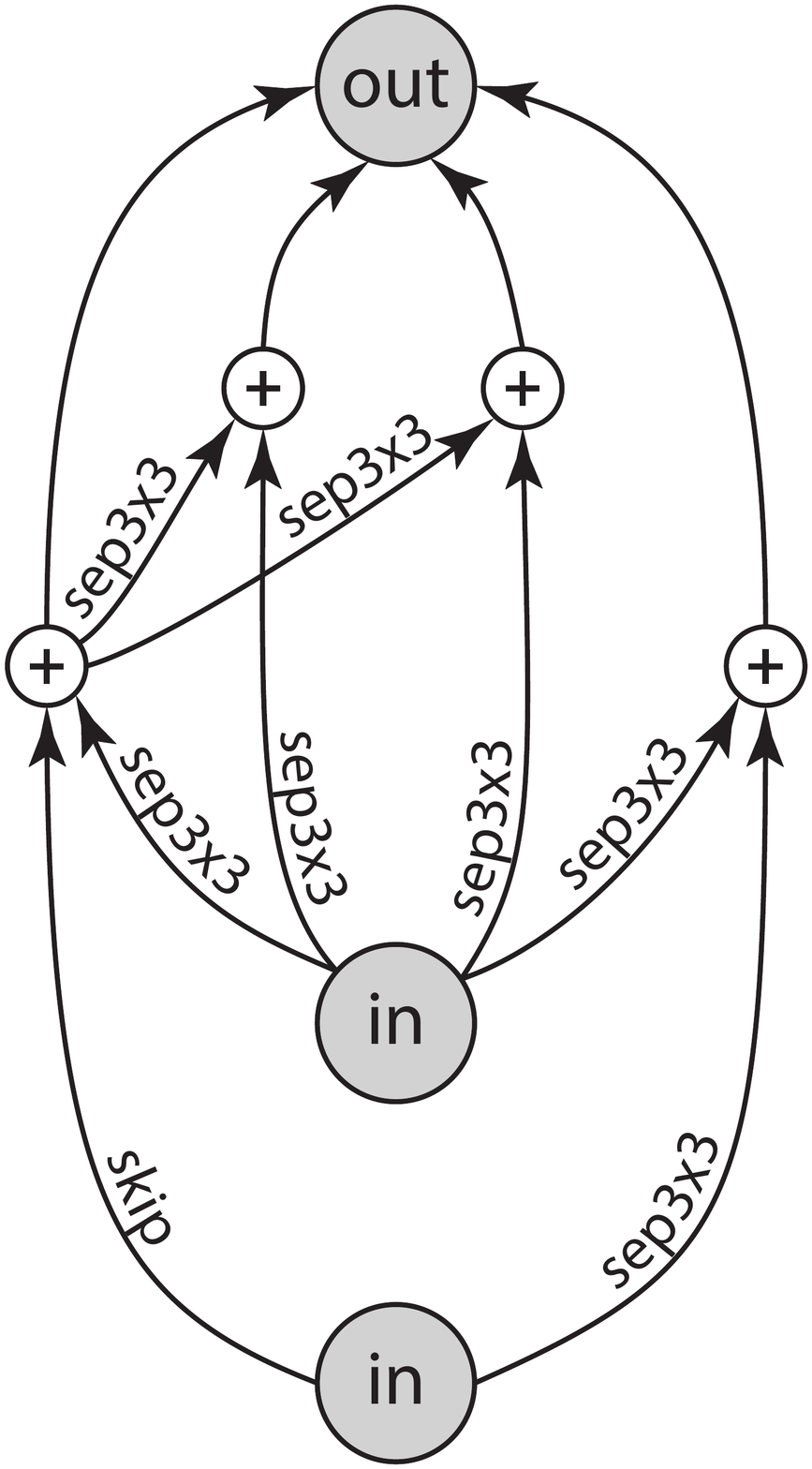}
\includegraphics[width=\figw]{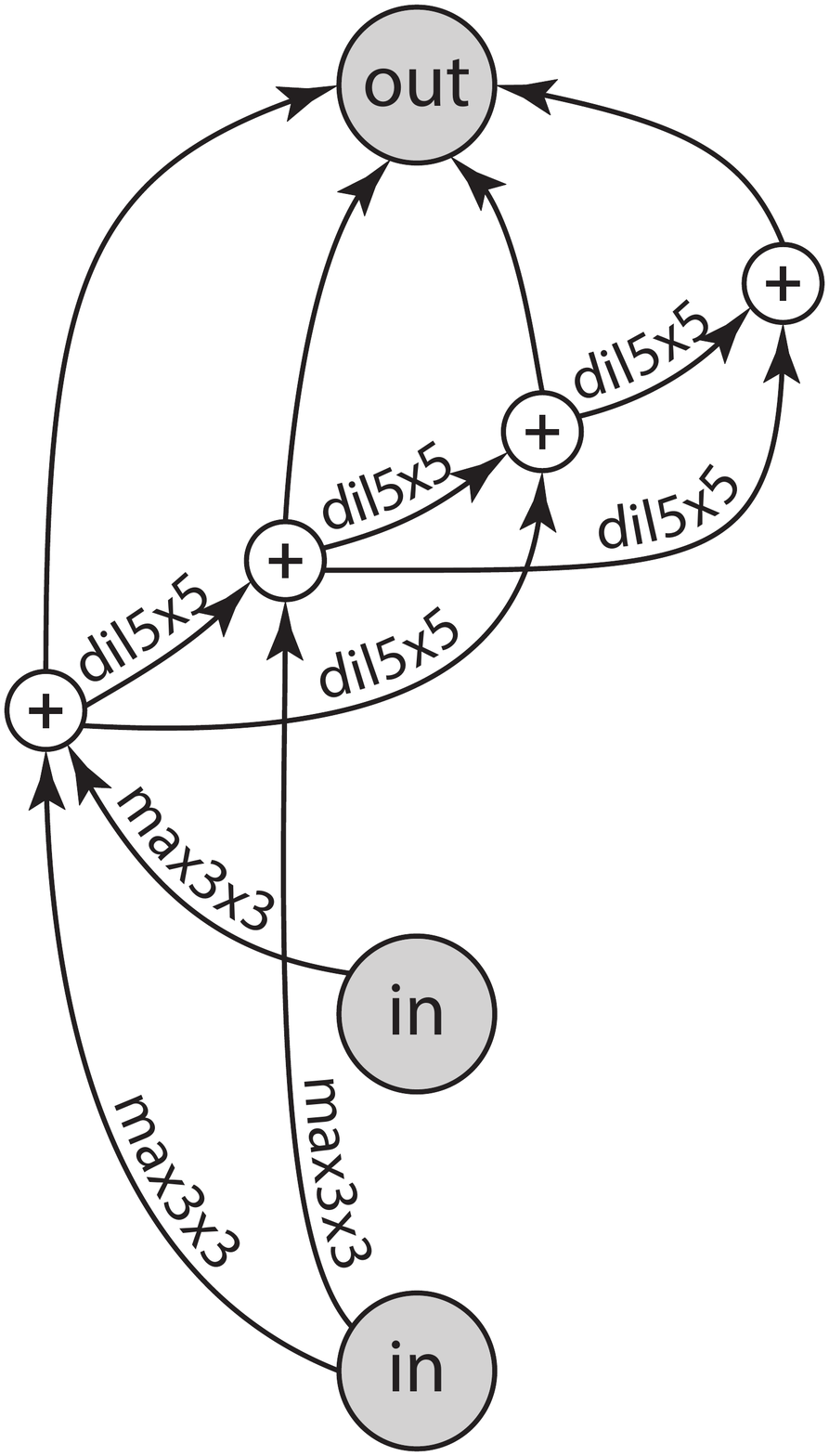}
\caption{NAS \cls{}}
\end{subfigure}
\begin{subfigure}[t]{0.24\textwidth}
\centering
\includegraphics[width=\figw]{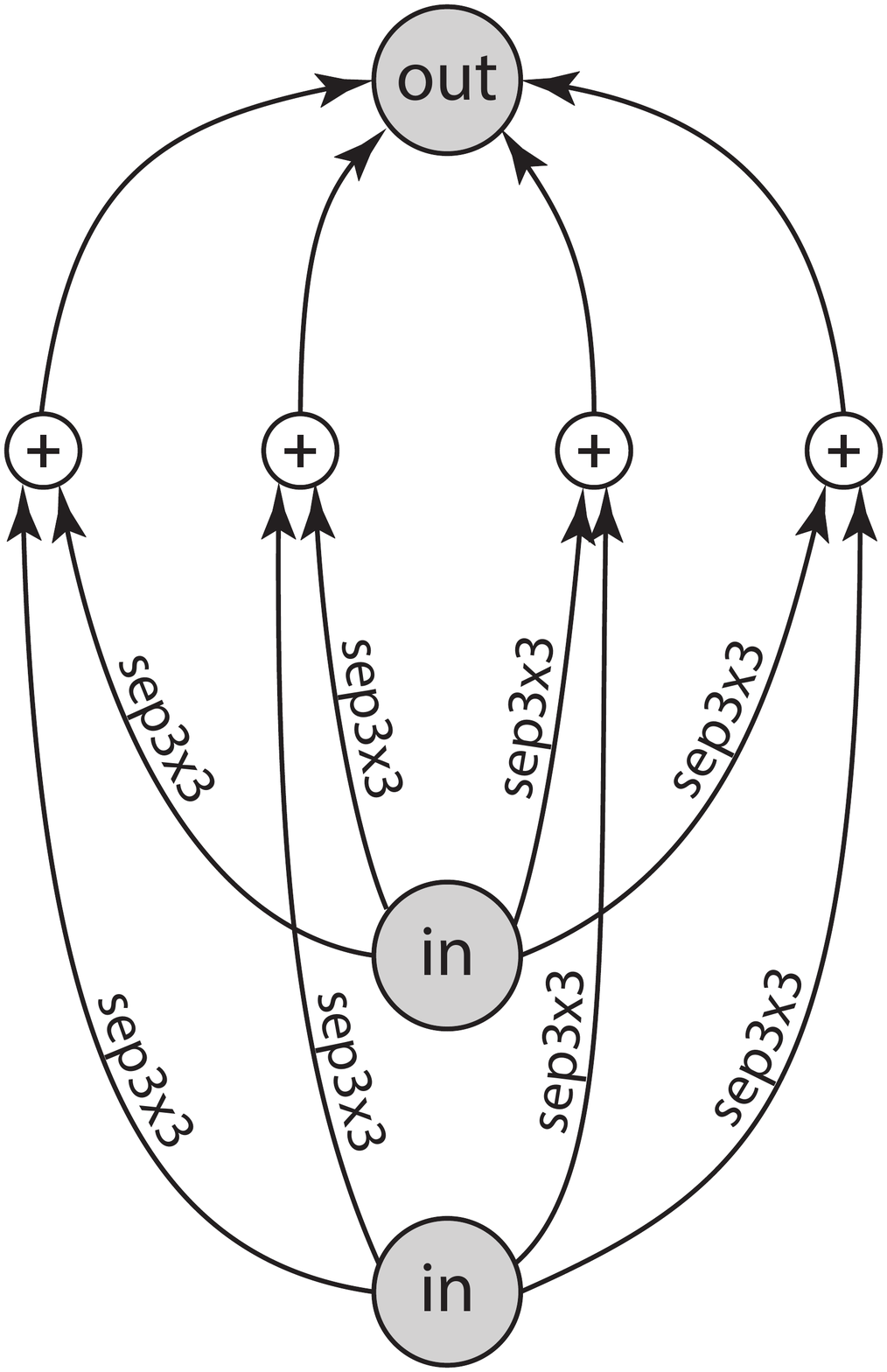}
\includegraphics[width=\figw]{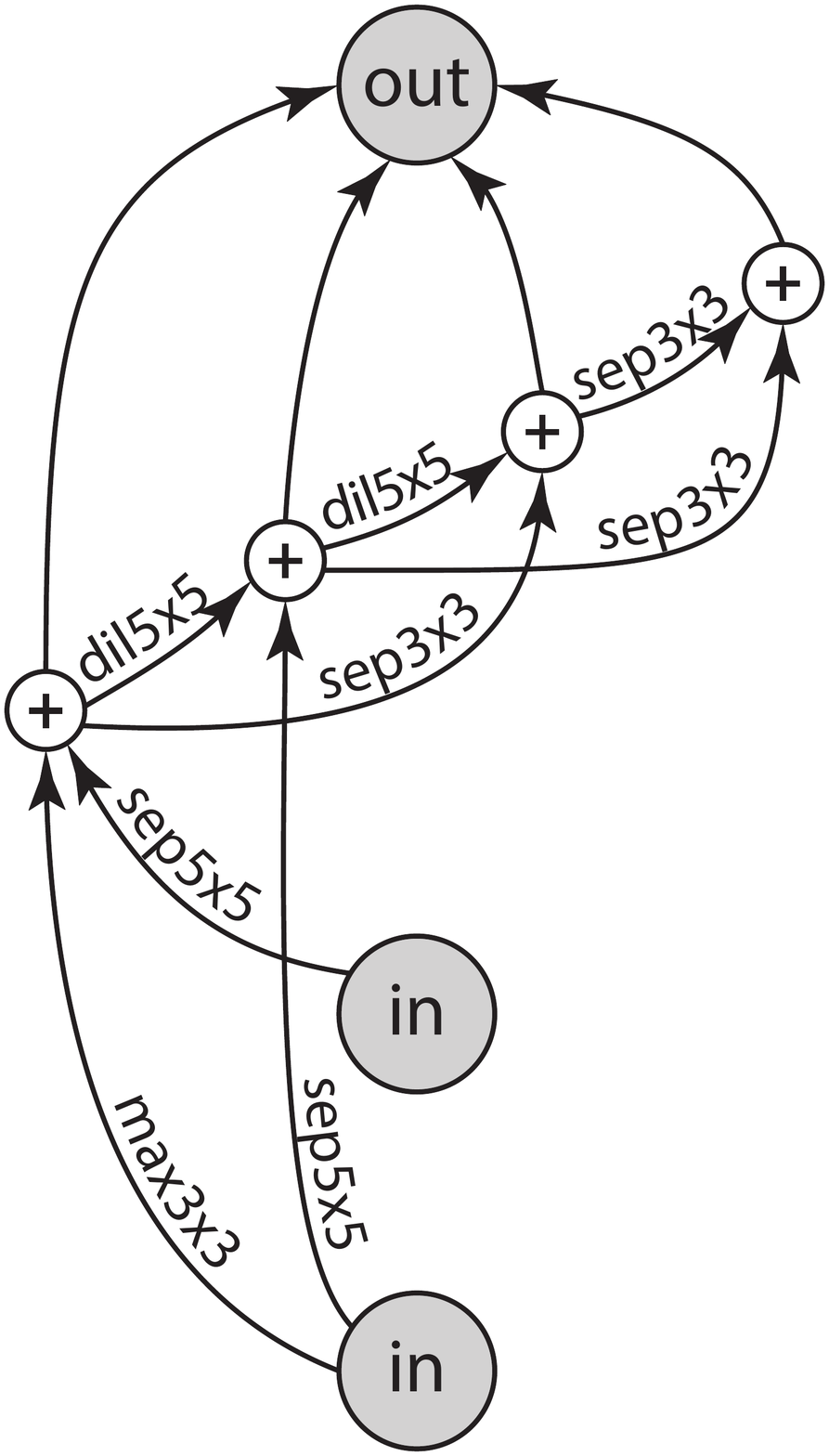}
\caption{\unnas{} \rot{}}
\end{subfigure}
\begin{subfigure}[t]{0.24\textwidth}
\centering
\includegraphics[width=\figw]{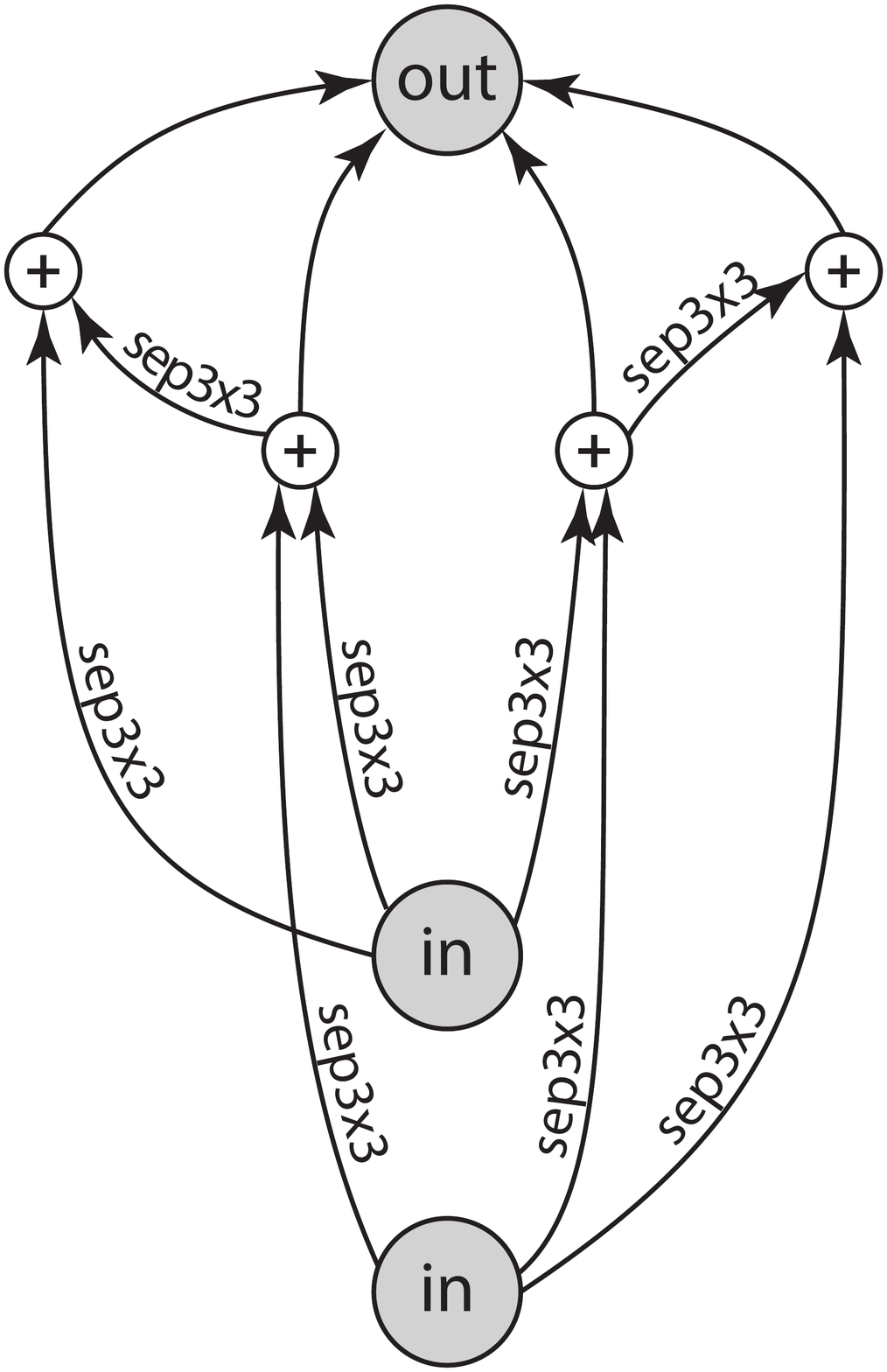}
\includegraphics[width=\figw]{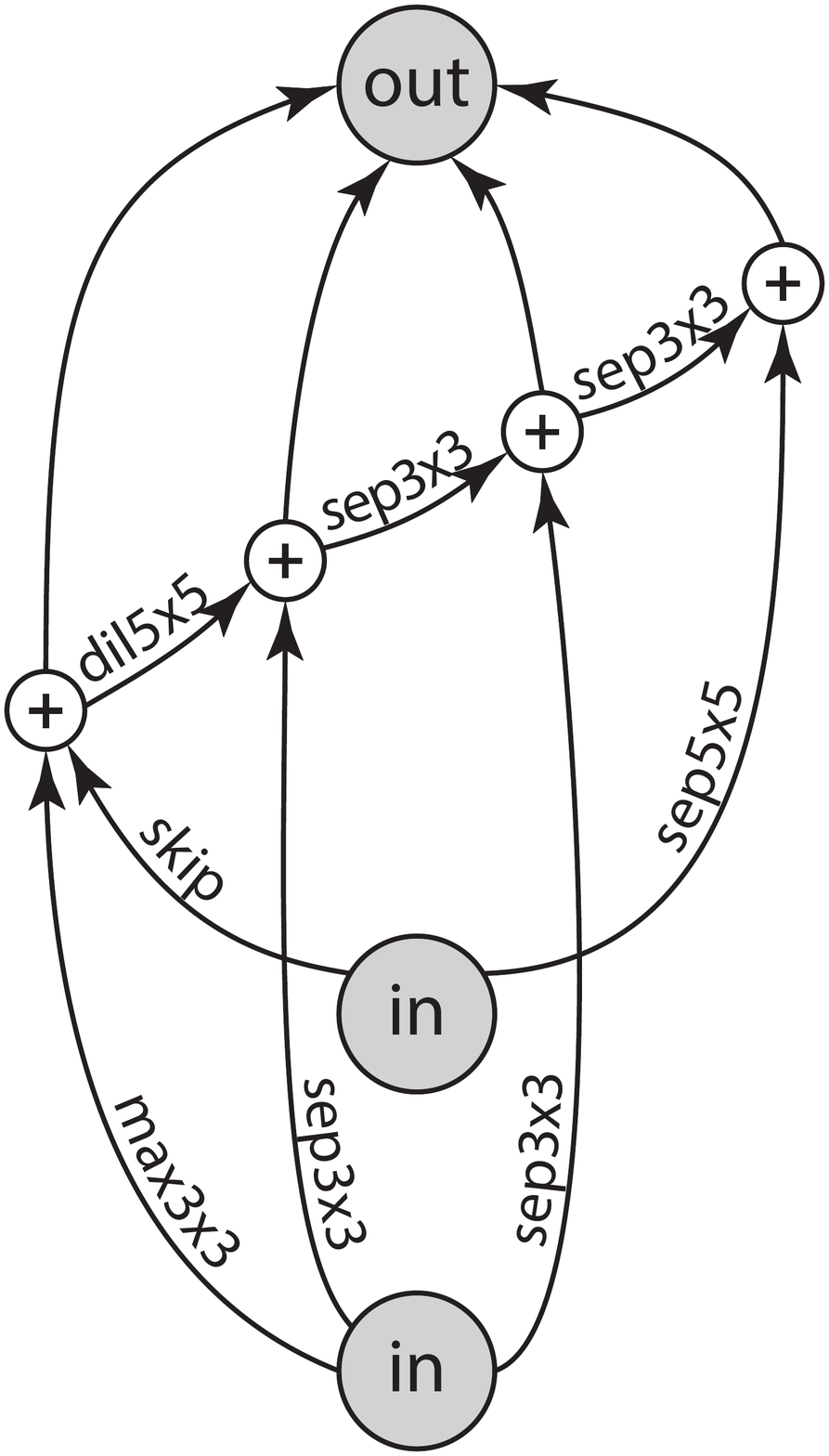}
\caption{\unnas{} \col{}}
\end{subfigure}
\begin{subfigure}[t]{0.24\textwidth}
\centering
\includegraphics[width=\figw]{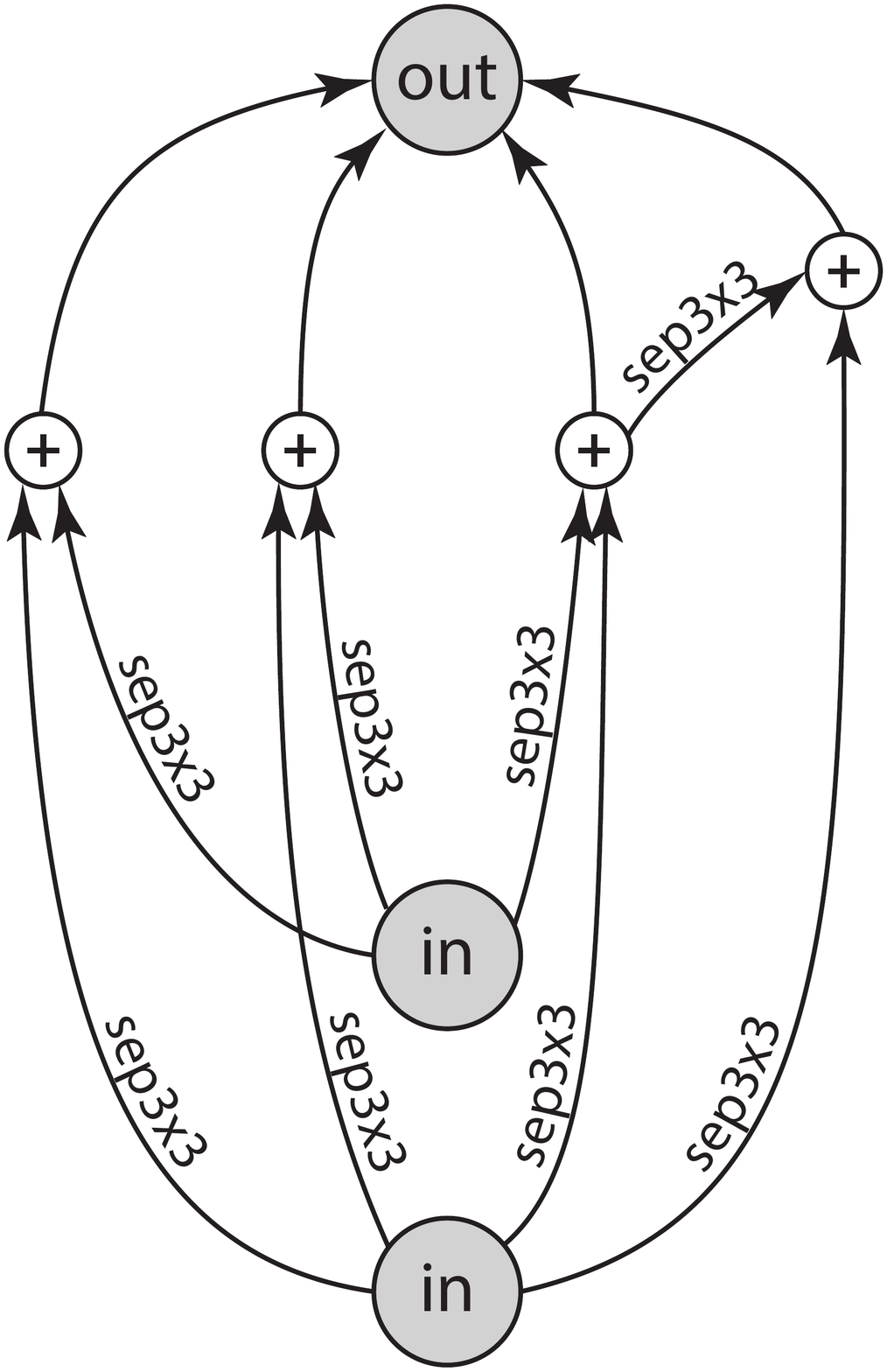}
\includegraphics[width=\figw]{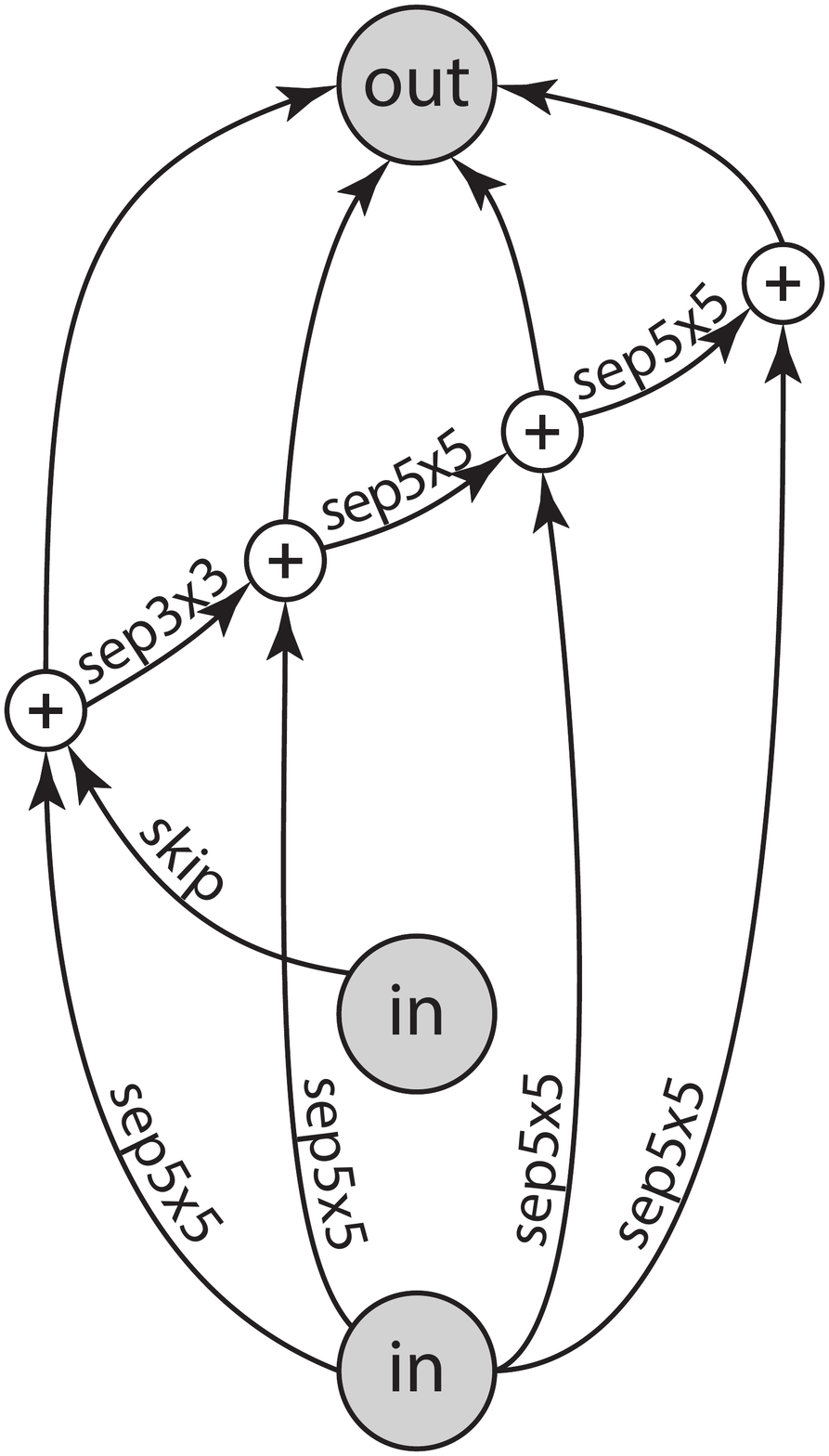}
\caption{\unnas{} \jig{}}
\end{subfigure}

\caption{Cell architectures (normal and reduce) searched on ImageNet-22K. }
\label{fig:darts_in22k_archs}
\end{figure}

\begin{figure}[!htp]\centering

\begin{subfigure}[t]{0.24\textwidth}
\centering
\includegraphics[width=\figw]{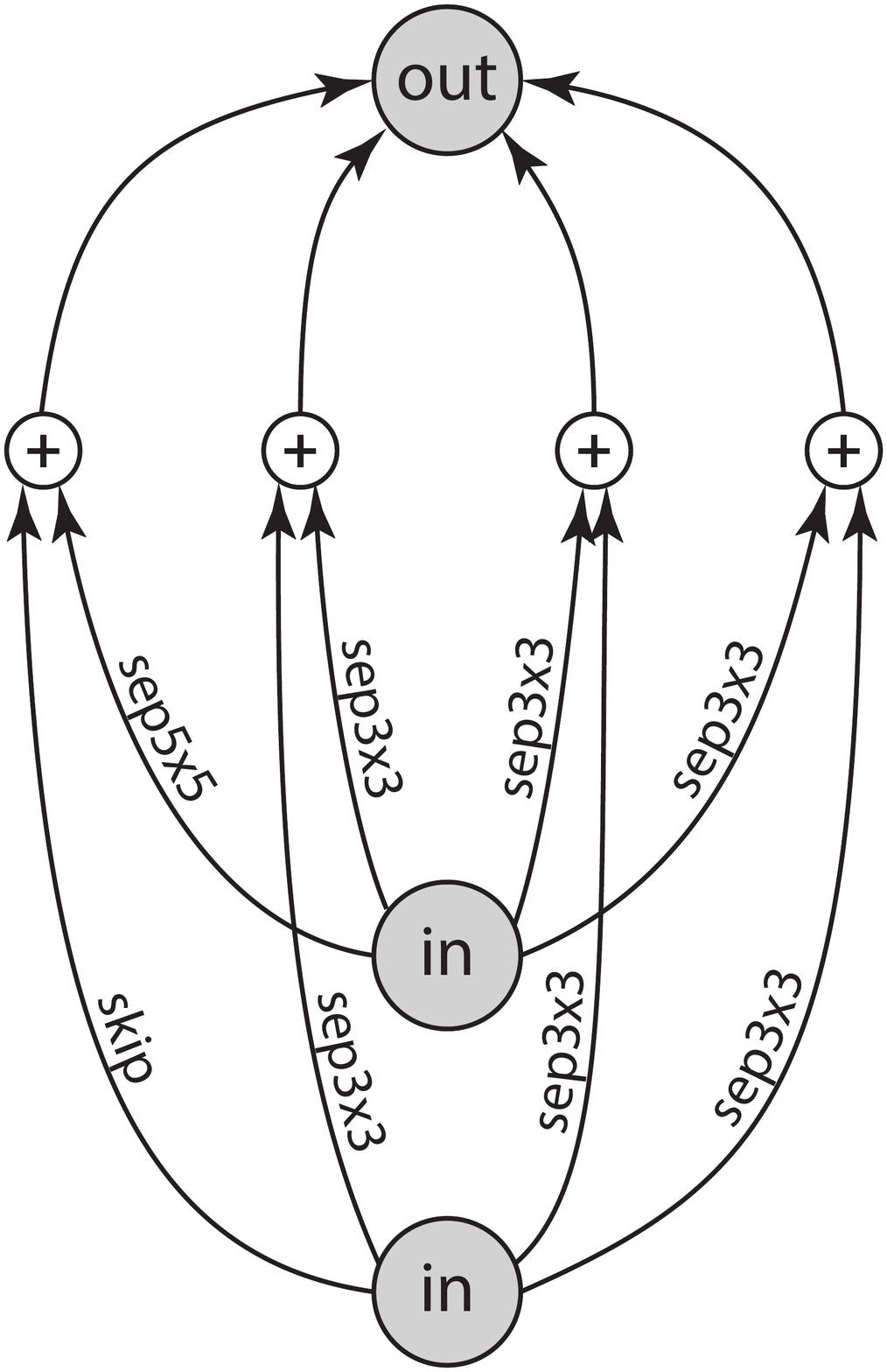}
\includegraphics[width=\figw]{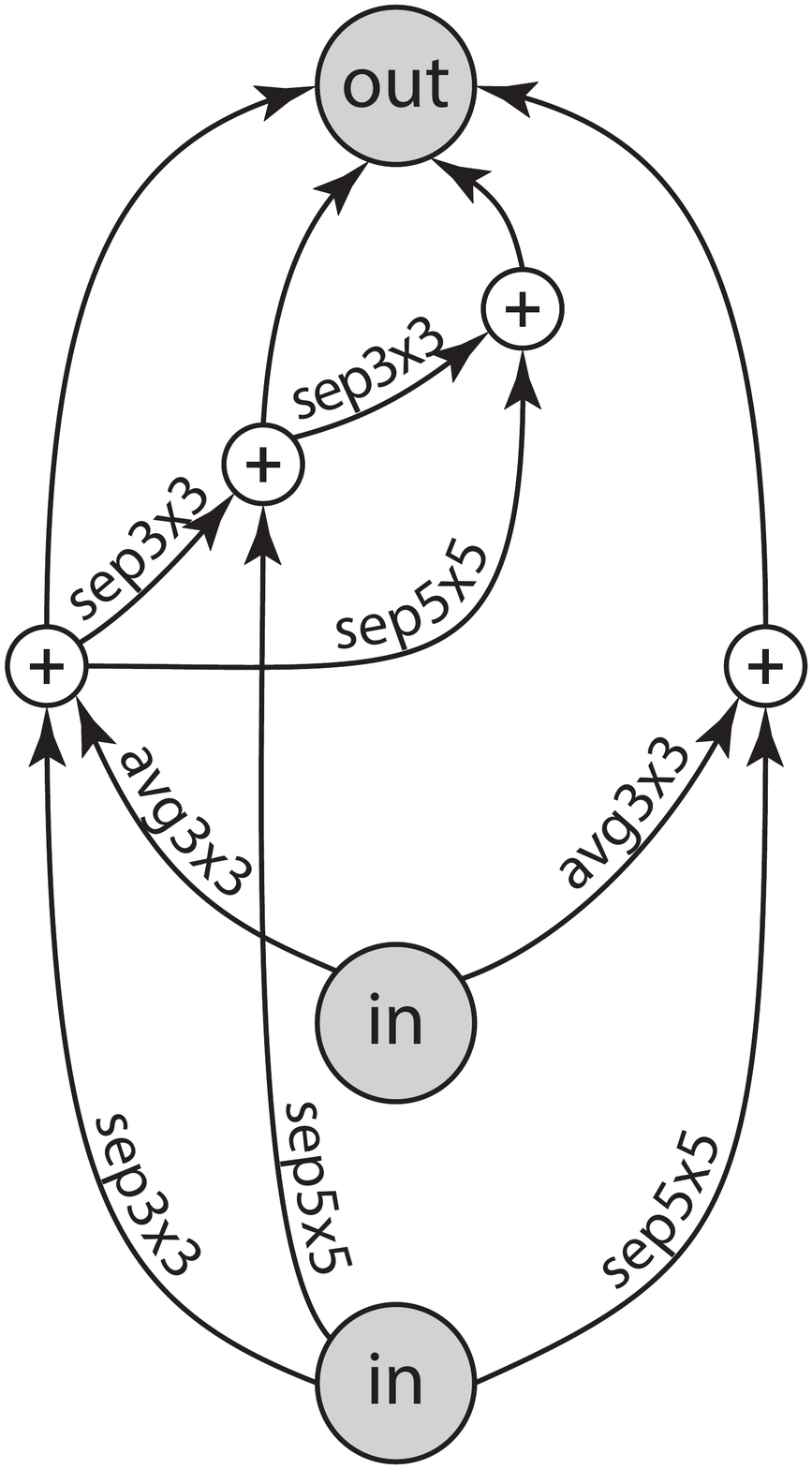}
\caption{NAS \seg{}}
\end{subfigure}
\begin{subfigure}[t]{0.24\textwidth}
\centering
\includegraphics[width=\figw]{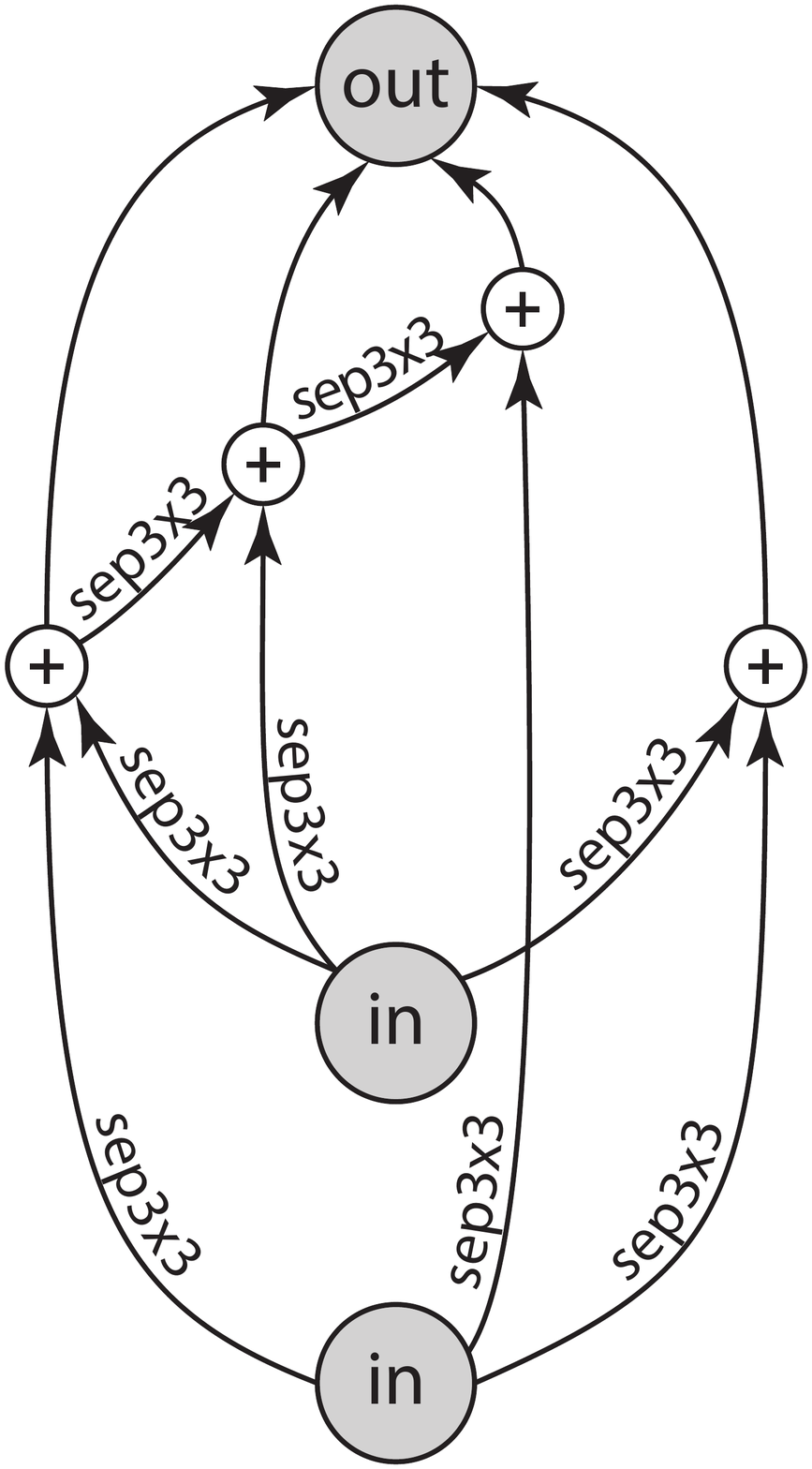}
\includegraphics[width=\figw]{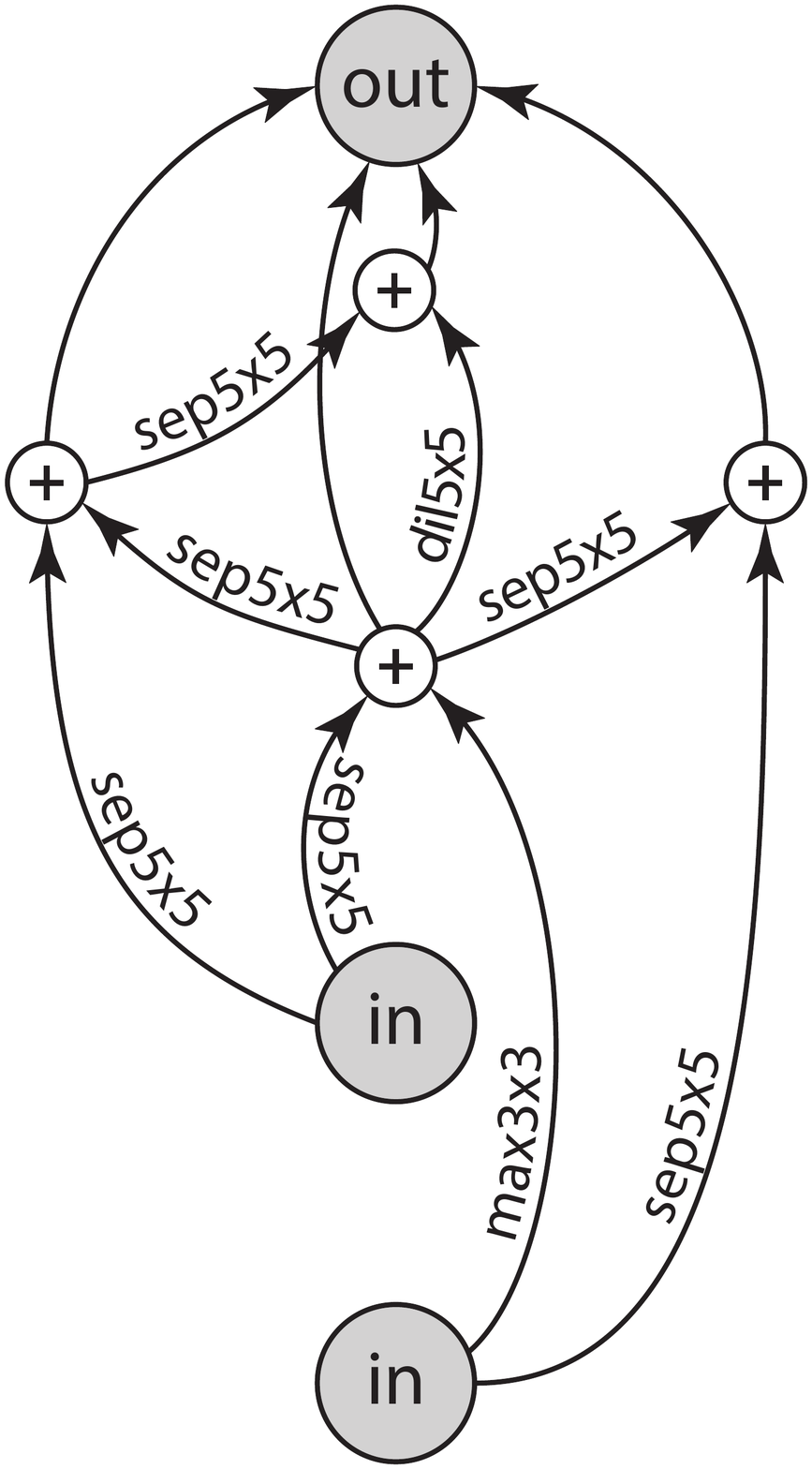}
\caption{\unnas{} \rot{}}
\end{subfigure}
\begin{subfigure}[t]{0.24\textwidth}
\centering
\includegraphics[width=\figw]{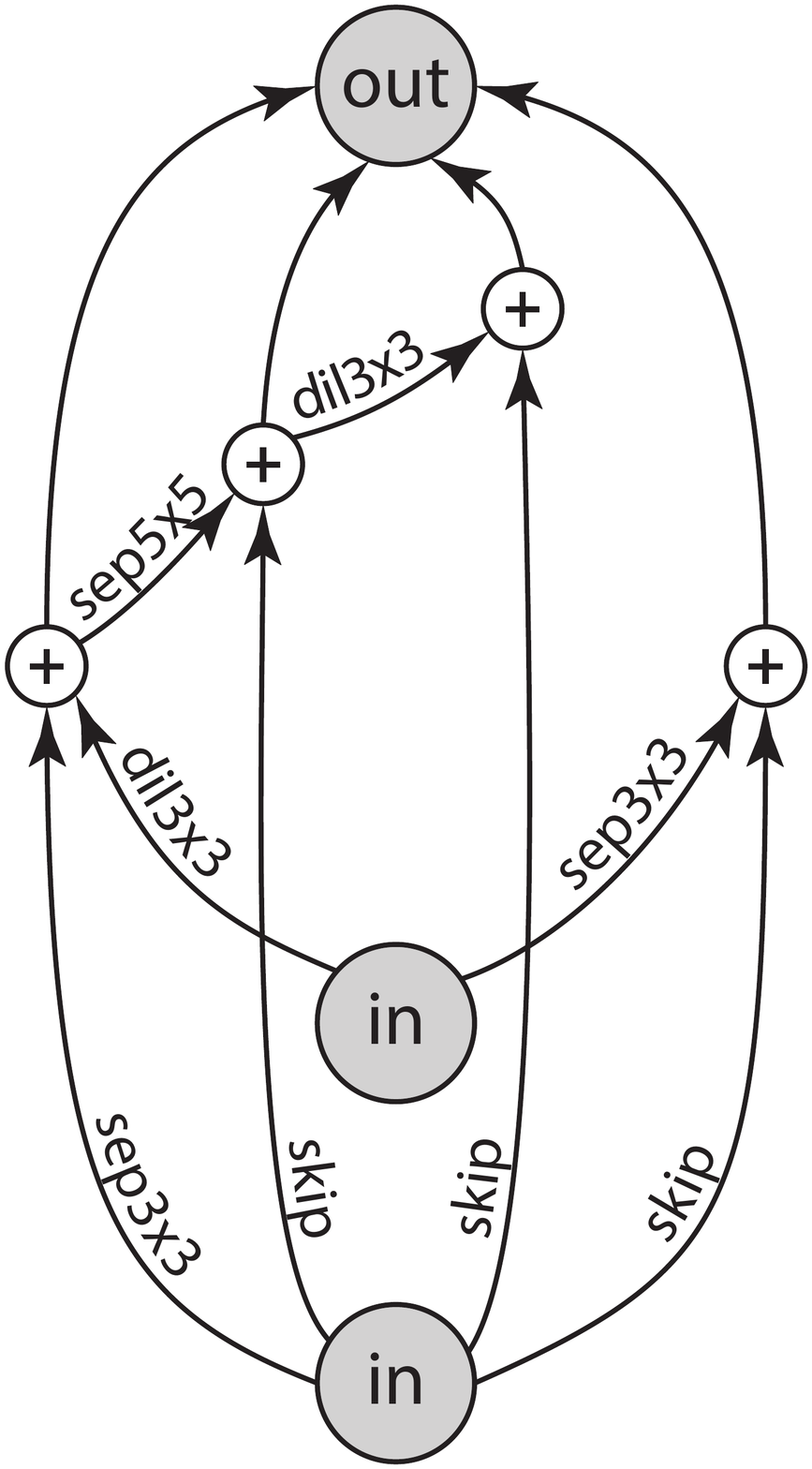}
\includegraphics[width=\figw]{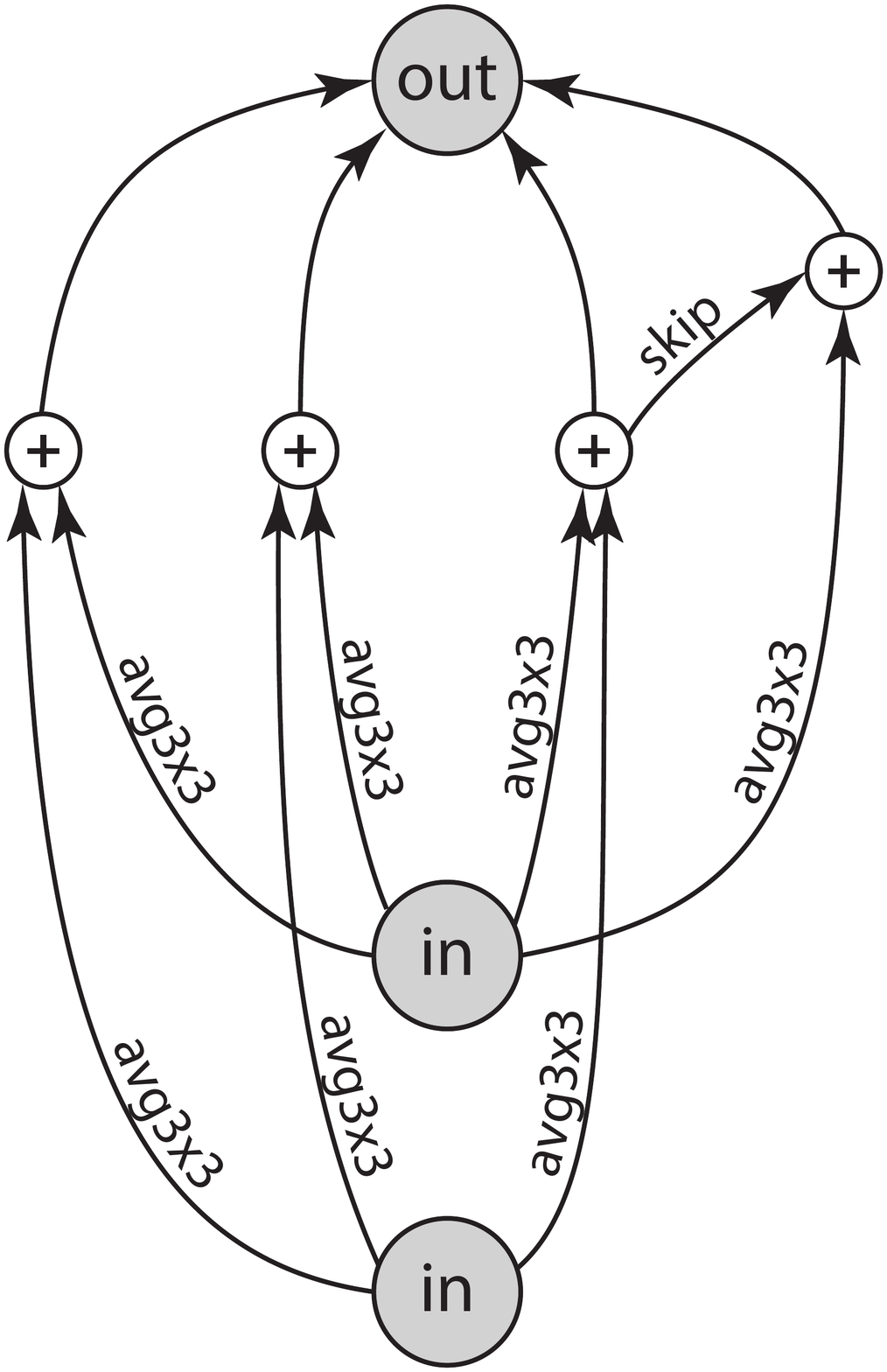}
\caption{\unnas{} \col{}}
\end{subfigure}
\begin{subfigure}[t]{0.24\textwidth}
\centering
\includegraphics[width=\figw]{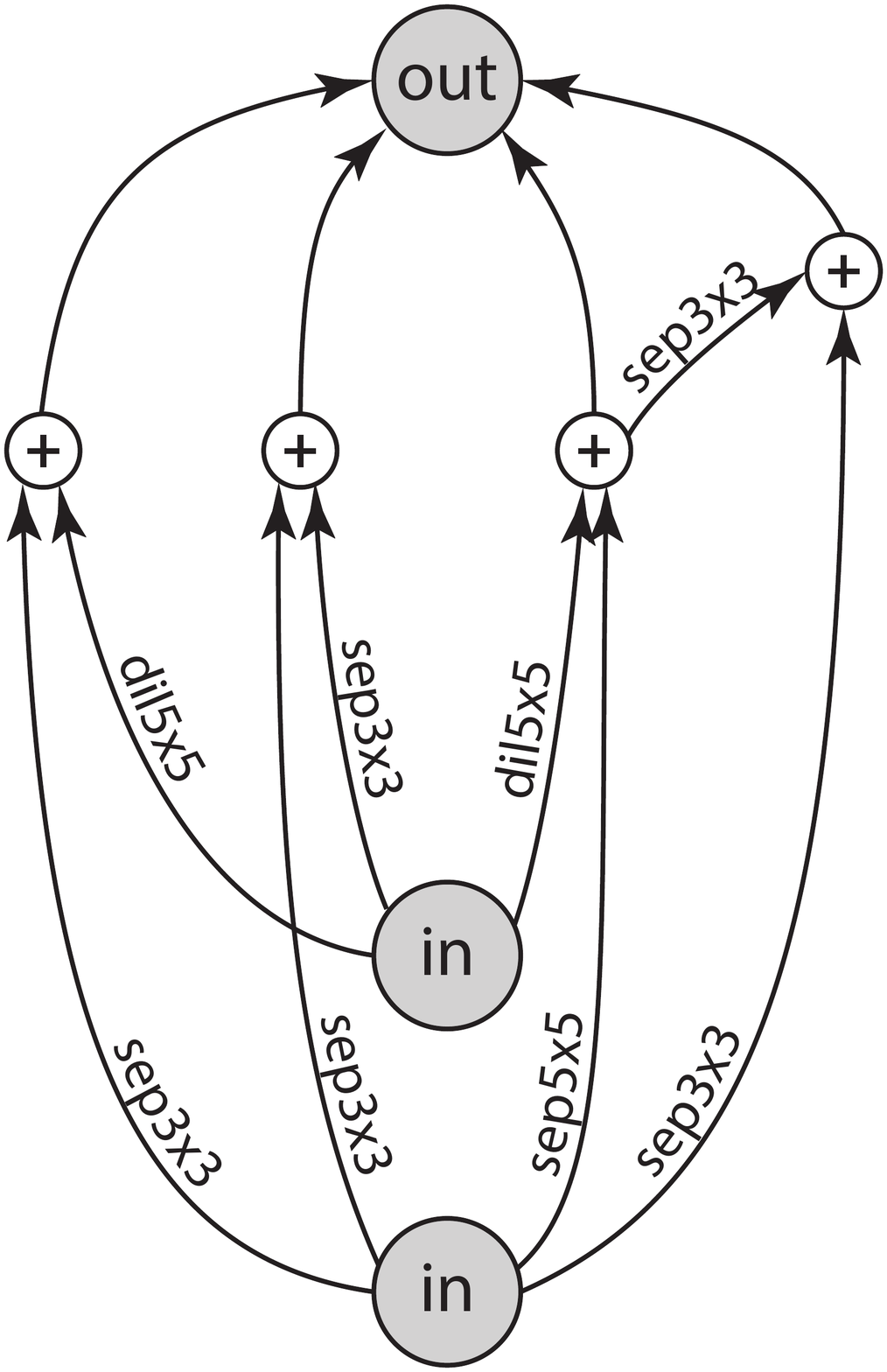}
\includegraphics[width=\figw]{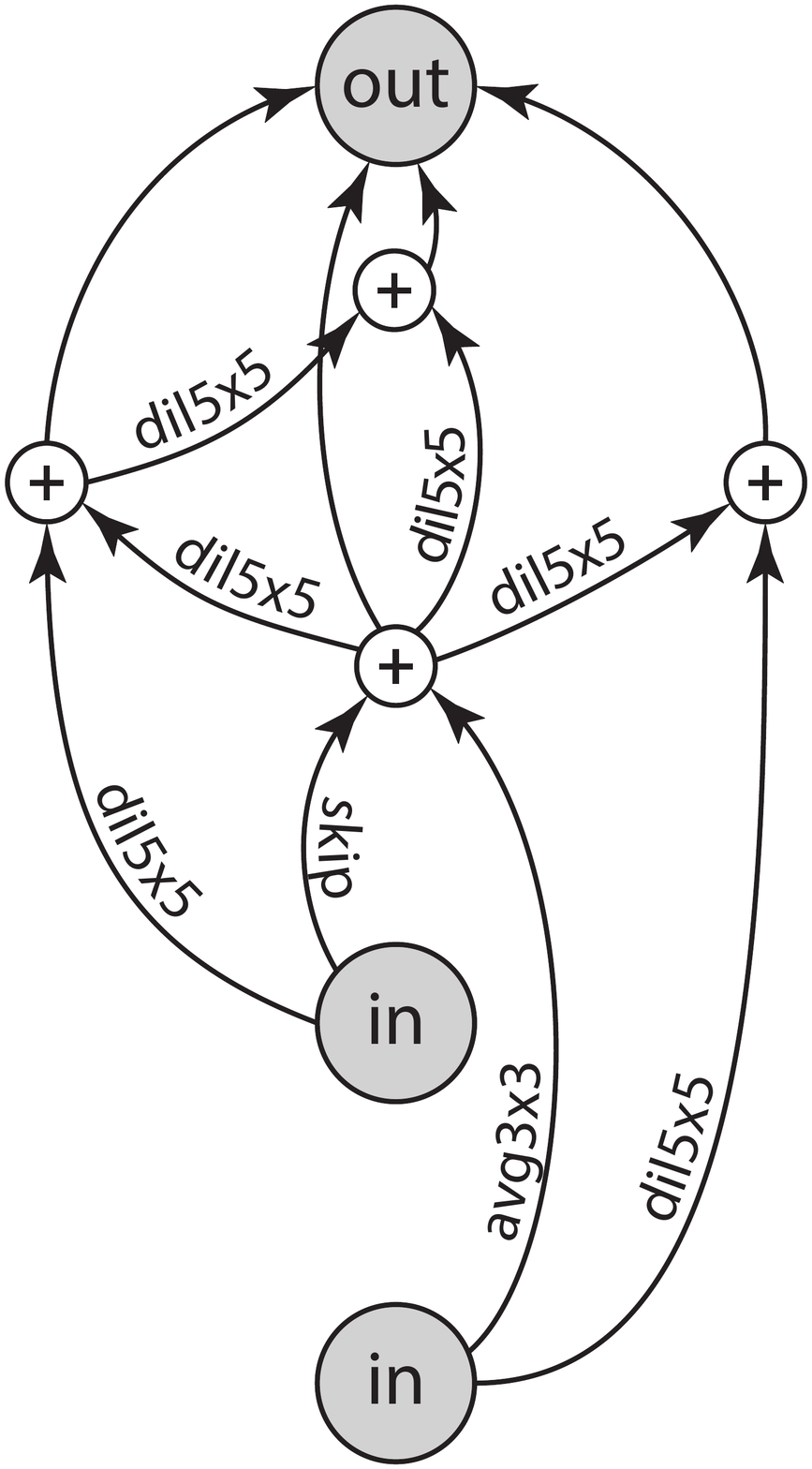}
\caption{\unnas{} \jig{}}
\end{subfigure}

\caption{Cell architectures (normal and reduce) searched on Cityscapes. }
\label{fig:darts_cityscapes_archs}
\end{figure}

\end{document}